\documentclass[lettersize,journal]{IEEEtran}
\usepackage{amsmath,amsfonts}
\usepackage{algorithm}
\usepackage{algorithmic}
\usepackage{array}
\usepackage[caption=false,font=normalsize,labelfont=sf,textfont=sf]{subfig}
\usepackage{textcomp}
\usepackage{stfloats}
\usepackage{url}
\usepackage{verbatim}
\usepackage{graphicx}
\usepackage{multirow}
\usepackage{xcolor}
\usepackage{booktabs}
\usepackage{diagbox}
\usepackage{pifont}
\usepackage{graphicx} 
\usepackage{booktabs}
\usepackage{amsmath}
\usepackage{multirow}
\usepackage{booktabs}
\usepackage{multirow}
\usepackage{makecell}
\usepackage{graphicx}
\usepackage{amssymb}
\usepackage{amsthm}
\usepackage{arydshln}
\usepackage{graphicx}
\usepackage{hyperref}
\usepackage{subcaption}
\usepackage{wrapfig}
\newtheorem{proposition}{Proposition}
\newtheorem{theorem}{Theorem}
\newtheorem{lemma}{Lemma}

\newtheorem{definition}{Definition}
\usepackage{pifont} 
\def\BibTeX{{\rm B\kern-.05em{\sc i\kern-.025em b}\kern-.08em
    T\kern-.1667em\lower.7ex\hbox{E}\kern-.125emX}}
\usepackage{balance}

\begin{document}
\title{SR$^2$-LoRA: Self-Rectifying Inter-layer Relations in Low-Rank Adaptation for Class-Incremental Learning}

\author{Fengqiang Wan, Yipeng Lin, Kan Lv, Yang Yang
\IEEEcompsocitemizethanks{\IEEEcompsocthanksitem Fengqiang Wan, Yipeng Lin, Kan Lv and Yang Yang are with the School of Computer Science and Engineering, Nanjing University of Science and Technology, Nanjing 210094, China. \protect\\
E-mail: \{fqwan, yipeng, lvkan, yyang\}@njust.edu.cn
\IEEEcompsocthanksitem Corresponding author: Yang~Yang.}
}

\maketitle

\begin{abstract}
Pre-trained models with parameter-efficient fine-tuning (PEFT) have demonstrated promising potential for class-incremental learning (CIL), yet catastrophic forgetting still persists when adapting models to new tasks. In this paper, we present a novel perspective on catastrophic forgetting through the analysis of inter-layer relation drift, i.e., the progressive disruption of relationships among layer-wise representations during the learning of new tasks. We theoretically show that the increase of such drift reduces the classification margins of previously learned tasks, thereby degrading overall model performance. To address this issue, we propose \underline{S}elf-\underline{R}ectifying inter-layer \underline{R}elation Low-Rank Adaptation~(SR$^2$-LoRA), a simple yet effective method that mitigates catastrophic forgetting by constraining inter-layer relation drift. Specifically, SR$^2$-LoRA constructs the relation matrices induced by the previous and current models on current-task samples, and aligns the corresponding singular values. We further theoretically show that this alignment exhibits greater robustness to estimation perturbations than direct entry-wise alignment. Extensive experiments on standard CIL benchmarks demonstrate that SR$^2$-LoRA effectively mitigates catastrophic forgetting, with its advantages becoming more pronounced as the number of tasks increases. Code is available in the \href{https://github.com/FqWan24/SR-2-LoRA}{repository}.

\end{abstract}

\begin{IEEEkeywords}
Class-Incremental Learning, Catastrophic Forgetting, Low-Rank Adaptation
\end{IEEEkeywords}

\section{Introduction}

\IEEEPARstart{C}{lass}-Incremental Learning (CIL) aims to enable models to sequentially incorporate new tasks while preserving performance on previously learned tasks~\cite{stability-plasticity-dilemma-2,de2021continual}, as required in real-world applications involving streaming or non-stationary data, such as visual recognition~\cite{DBLP:journals/nature/SilverHMGSDSAPL16, 2021Highly} and autonomous systems~\cite{Kasabov2005IncrementalLI, DBLP:journals/jirs/ShaheenHHS22, DBLP:conf/avss/ZaalICMR19}. However, such sequential adaptation inevitably induces catastrophic forgetting, wherein performance on earlier tasks progressively deteriorates as new tasks are incorporated~\cite{french1999catastrophic,french2020modeling,cf_1_kirkpatrick2017overcoming}. Most existing approaches adopt a learning-from-scratch paradigm~\cite{DBLP:conf/cvpr/wa,DBLP:conf/iclr/memo,zhou2023pycil}, which often suffers from limited generalization~\cite{zhang2023lora,vpt:conf/nips/LuZCX00Z24}. In contrast, pre-trained models (PTMs) exhibit strong transferability capabilities, making them more suitable for CIL~\cite{qiu2020pre,CLIP,steiner2021train}. Building on this advantage, recent studies have increasingly turned to parameter-efficient fine-tuning (PEFT) to adapt PTMs while alleviating catastrophic forgetting~\cite{liang2024inflora,wang2022dualprompt,zhou2024expandable}.

Existing methods can be broadly categorized into selection-based methods~\cite{wang2022learning}, prototype-based methods~\cite{learn_drift:conf/eccv/GomezVillaGWBTW24,theory_proto:conf/icml/WangR000BN25}, covariance-based methods~\cite{calibration:conf/cvpr/ChenDWHGL25,semantic_drift:conf/icml/He0XC00ZZ25}, and orthogonality-based methods~\cite{prompt_oloss:conf/aaai/LuZC0X0025,o_loss:conf/iccv/ShiY23,bilora:conf/cvpr/ZhuZDK25}. Selection-based methods~\cite{smith2023coda,wang2022dualprompt} attribute catastrophic forgetting to incorrect module selection and mitigate it by improving out-of-distribution (OOD) detection~\cite{OE_2018,hendrycks2021many}, thereby reducing overconfident predictions from mismatched modules. Prototype-based methods~\cite{acmap_2025,noise_mix:journals/corr/abs-2509-16738,zhou2024expandable} associate catastrophic forgetting with shifts in class prototypes and estimate such shifts using current-task samples to correct prototype representations, thereby alleviating cross-task misalignment. Covariance-based methods~\cite{calibration_2:journals/corr/abs-2511-10974,calibration_3:journals/corr/abs-2511-09926,wu2025navigating} further model catastrophic forgetting as covariance drift and adjust class-wise covariance statistics to preserve previously learned feature distributions. In contrast, orthogonality-based methods~\cite{liang2024inflora,split_lora:journals/corr/abs-2505-22370,tuna_2025} attribute catastrophic forgetting to gradient interference across tasks and mitigate it by projecting current-task gradients onto subspaces orthogonal to previous tasks, thereby reducing interference with prior knowledge.

\begin{figure}[t]
  \centering
  \includegraphics[width=0.9\linewidth]{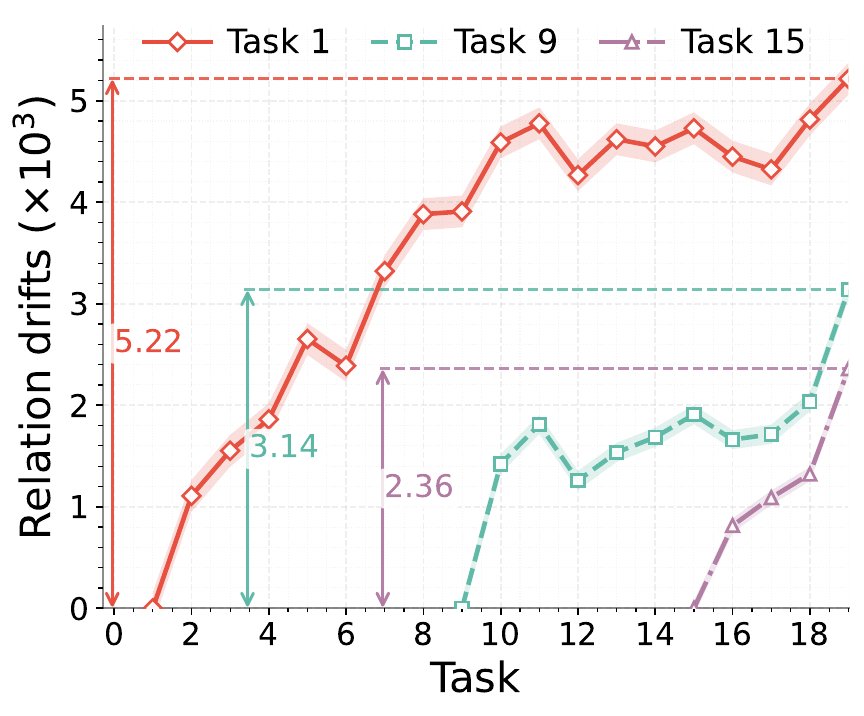}
  \caption{Inter-layer relation drift during incremental training. For each selected task, the drift is measured on that task's samples by comparing the inter-layer relation induced by the model obtained immediately after learning that task and by the subsequent models learned in later tasks. Results on ImageNet-R with 20 task (illustrated for Tasks 1, 9, and 15) show that the drift generally increases as more tasks are learned, suggesting that inter-layer relations are progressively disrupted throughout the incremental learning process.}
  \label{fig:intro_drift}
\end{figure}

In this paper, we examine catastrophic forgetting from the perspective of \emph{inter-layer relation drift}, where inter-layer relations refer to the relationships among layer-wise representations within a model. As shown in Fig~\ref{fig:intro_drift}, for a given task, inter-layer relation drift is measured on its samples by comparing the representations produced by the model learned on that task and by the subsequent models, and it increases as more tasks are learned. This observation indicates that learning new tasks progressively disrupts the inter-layer relations established for previous tasks. Since the knowledge of previous tasks is encoded not only in layer-wise representations but also in inter-layer relations~\cite{layer_classifier,kong2025cr}, increasing inter-layer relation drift reduces the classification margins of previous tasks and thereby degrades model performance, as formalized in Theorem~\ref{theorem_forgetting_relation}. These results suggest that constraining inter-layer relation drift constitutes a principled approach to mitigating catastrophic forgetting.

The above analysis naturally motivates our proposed Self-Rectifying inter-layer Relations Low-Rank Adaptation~(SR$^2$-LoRA), a simple yet effective method for mitigating catastrophic forgetting by constraining inter-layer relation drift. Specifically, for each current-task sample, SR$^2$-LoRA constructs the inter-layer relation matrices induced by the model learned on previous tasks and the current model, respectively. Singular value decomposition (SVD)~\cite{svd:journals/eswa/SunPLC26} is applied to both matrices, and the corresponding singular values are aligned during training. Since the relations induced by the previous model on current-task samples are subject to estimation errors, direct entry-wise alignment is sensitive to such perturbations. In contrast, singular value alignment exhibits improved robustness to such perturbations, consistent with the stability guarantee established in Proposition~\ref{prop:sv_stability}. Extensive experiments on standard CIL benchmarks~\cite{zhou2024expandable,wu2025navigating,sun2025mos} demonstrate that SR$^2$-LoRA effectively mitigates catastrophic forgetting, with performance gains becoming increasingly pronounced as the number of tasks increases.

The contributions of this work are summarized as follows:

\begin{itemize}
    \item We provide a theoretical perspective on catastrophic forgetting from the viewpoint of inter-layer relation drift. We establish that increasing inter-layer relation drift reduces the classification margins of previous tasks, thereby leading to performance degradation.

    \item Motivated by this analysis, we propose Self-Rectifying inter-layer Relation Low-Rank Adaptation, a simple yet effective method that mitigates catastrophic forgetting by constraining inter-layer relation drift via singular value alignment. We further show that singular value alignment exhibits improved robustness to estimation perturbations compared with direct entry-wise alignment.

    \item We conduct extensive experiments on CIL benchmarks, demonstrating that SR$^2$-LoRA effectively mitigates catastrophic forgetting, with performance gains becoming increasingly pronounced as the number of tasks increases.
\end{itemize}

\section{Related Work}

\subsection{PTMs-Based Class-Incremental Learning}

The advent of large-scale pre-trained models (PTMs)~\cite{steiner2021train,CLIP} has significantly advanced downstream learning. Consequently, parameter-efficient fine-tuning (PEFT)~\cite{han2024parameter,houlsby2019parameter} has emerged as a prevalent paradigm for class-incremental learning. Existing methods can be broadly grouped into four categories according to their forgetting mitigation strategies, including selection-based, prototype-based, covariance-based, and orthogonality-based methods. Selection-based methods mitigate forgetting by improving module selection or composition across tasks. CODA-Prompt~\cite{smith2023coda} composes prompts through a decomposed attention mechanism, while MOS~\cite{sun2025mos} dynamically merges adapters via a self-optimization retrieval strategy. CL-LoRA~\cite{he2025cl} further selects task-specific adaptation modules using learnable block-wise weights. Prototype-based methods address forgetting by rectifying prototype drift across tasks. SSIAT~\cite{tan2024semantically} aligns features of previous and current tasks through shared adapter tuning and prototype shift estimation. LoRA-DRS~\cite{liu2025lora} reduces representation drift by constructing a drift-resistant feature space that removes task-specific variations. EASE~\cite{zhou2024expandable} mitigates prototype drift by integrating multiple adapter predictions with semantic-guided prototype synthesis. ACMap~\cite{acmap_2025} aligns previous prototypes within a consolidated adapter subspace via centroid mapping to correct feature drift. Covariance-based methods mitigate forgetting by modeling distributional shifts through covariance calibration. MACIL~\cite{wu2025navigating} aligns feature distributions across tasks via mean-shift compensation and covariance calibration. SLDC~\cite{calibration_3:journals/corr/abs-2511-09926} mitigates distribution drift by learning a latent transition operator that aligns Gaussian distributions of previous tasks with the current model. Orthogonality-based methods mitigate forgetting by reducing gradient interference across tasks. InfLoRA~\cite{liang2024inflora} enforces orthogonality constraints to isolate low-rank subspaces, while BiLoRA~\cite{bilora:conf/cvpr/ZhuZDK25} expands the effective parameter space through a bilinear formulation to support a larger number of tasks with limited interference.

\subsection{Knowledge Distillation}
Knowledge distillation~\cite{logtis_1:conf/aaai/GaoHZXZMDW25, feature_1:conf/cvpr/KangPH22} is a widely adopted learning paradigm in which a compact student model is trained to mimic the behavior of a more powerful teacher model. The concept was first introduced in~\cite{DBLP:conf/kdd/BucilaCN06} and subsequently formalized and popularized in~\cite{DBLP:journals/corr/HintonVD15}. Depending on the form of transferred knowledge, existing approaches can be broadly categorized into logits-based, feature-based, and relation-based methods. Logits-based methods~\cite{logtis_2:conf/cvpr/SunR00C24,logtis3:journals/tim/LiXYHSSY25} transfer class-level dark knowledge by aligning the softened output distributions of teacher and student models, thereby encouraging the student to approximate the predictive behavior of the teacher. Feature-based methods~\cite{feature_2:conf/iccvw/JodeletLPM23,feature_3:journals/access/KhanBR24} guide the student model by matching intermediate representations, which promotes the learning of more discriminative and transferable feature embeddings. Relation-based methods~\cite{relation_1,relation_2,relation_3,relation_4,relation_5} focus on preserving structural relationships among samples, enabling the student to capture relational inductive biases that go beyond individual predictions and improve generalization. Self-distillation can be viewed as a special case of knowledge distillation in which the teacher and student share the same architecture. In this setting, knowledge is transferred within a single model to progressively refine its internal representations across layers or training stages. For instance,~\cite{self_distill} distill knowledge from deeper layers to shallower ones, while~\cite{self_dis_2} transfer information from early-stage representations to later layers, facilitating more effective representation learning.

\section{Method}
\label{sec:method}
\begin{figure*}[t]
  \begin{center}
    \centerline{\includegraphics[width=\textwidth]{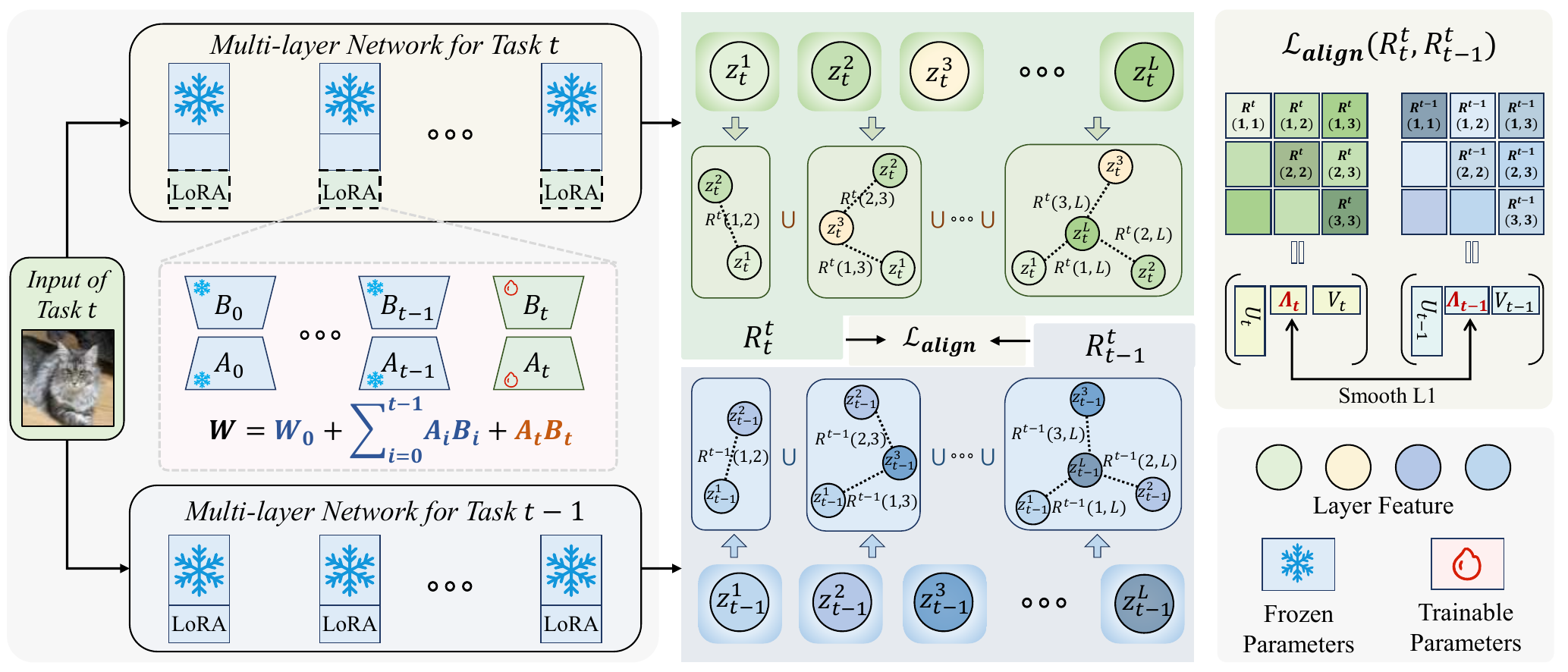}}
    \caption{Illustration of the proposed framework. During training of task \(t\), the backbone \(W_0\) and all previously learned LoRA modules \(\sum_{i=1}^{t-1} B_i A_i\) are frozen, while only the newly introduced module \(B_t A_t\) is optimized. A constraint on the drift of inter-layer relations is imposed between the relation matrix \(R_{t-1}\) induced by the previous model and the relation matrix \(R_t\) induced by the current model on samples from task \(t\). To achieve this, singular value decomposition (SVD)~\cite{svd:journals/eswa/SunPLC26} is applied to both relation matrices, and alignment is performed solely on their singular values.}
    \label{method:overview}
  \end{center}
\end{figure*}

\subsection{Preliminaries}

We consider CIL with a sequence of tasks indexed by \(t \in \{1,\dots,T\}\). At task \(t\), the model observes a dataset \(\mathcal{D}_t=\{(x_t^i,y_t^i)\}_{i=1}^{n_t}\), where \(x_t^i\in\mathcal{X}\) denotes an input sample and \(y_t^i\in\mathcal{Y}_t\) denotes its corresponding label. The label sets are mutually disjoint, i.e., \(\mathcal{Y}_t \cap \mathcal{Y}_{t'}=\varnothing\) for \(t\neq t'\). We further denote by \(\mathcal{C}_t=\bigcup_{\tau=1}^{t}\mathcal{Y}_\tau\) the cumulative label space after learning task \(t\).

\textbf{Parameterization and Training Objective.}
Let \(W_0=\{W_0^\ell\}_{\ell=1}^{L}\) denote an \(L\)-layer pre-trained model. When learning task \(t\), the pre-trained parameters and all previously learned low-rank modules \(\{B_j^\ell A_j^\ell\}_{j=1}^{t-1}\) are fixed, and a new trainable low-rank branch \(\Delta W_t^\ell=B_t^\ell A_t^\ell\) is introduced at each layer \(\ell\). Here, \(A_t^\ell\in\mathbb{R}^{r\times d}\) is initialized from a Gaussian distribution, and \(B_t^\ell\in\mathbb{R}^{d\times r}\) is initialized to zero. The effective weight at layer \(\ell\) is given by
\begin{equation}
W_t^\ell
= W_{t-1}^\ell + B_t^\ell A_t^\ell
= W_0^\ell + \sum_{j=1}^{t} B_j^\ell A_j^\ell.
\end{equation}
We denote by \(W_t=\{W_t^\ell\}_{\ell=1}^{L}\) the model parameters at task \(t\). Given an input sample \(x_t^i\), let \(z_{t,t}^{0,i}\) denote its input embedding under the model \(W_t\), where the first subscript indicates the task index of the sample and the second denotes the model at task \(t\). For each layer \(\ell\in\{1,\ldots,L\}\), the hidden representation is defined recursively as \(z_{t,t}^{\ell,i} = z_{t,t}^{\ell-1,i} + h_{t,t}^{\ell,i}\), where \(h_{t,t}^{\ell,i} = \sigma\!\left(f(z_{t,t}^{\ell-1,i}; W_t^\ell)\right)\) denotes the residual branch output at layer \(\ell\). Here, \(\sigma(\cdot)\) is a nonlinear activation function and \(f(\cdot;W_t^\ell)\) denotes the layer-wise transformation parameterized by \(W_t^\ell\). The final representation is given by \(z_{t,t}^{L,i} = z_{t,t}^{0,i} + \sum_{\ell=1}^{L} h_{t,t}^{\ell,i}\).

During training on task \(t\), the model is optimized using the cross-entropy loss~\cite{logtisnorm:conf/icml/WeiXCF0L22}
\begin{equation}
\mathcal{L}_{\mathrm{CE}}
=
-\frac{1}{n_t}
\sum_{i=1}^{n_t}
\log
\frac{\exp\!\left((z_{t,t}^{L,i})^\top w_{y_t^i}\right)}
{\sum_{k \in \mathcal{Y}_t} \exp\!\left((z_{t,t}^{L,i})^\top w_k\right)},
\label{eq:ce}
\end{equation}
where \(w_k\) denotes the classifier weight corresponding to class \(k\). After completing task \(t\), the classifier over all observed classes \(W_t^c = [\,w_k\,]_{k \in \mathcal{C}_t}\) is recalibrated using pseudo-features sampled from class-wise Gaussian distributions estimated from all learned classes~\cite{zhang2023slca}. Specifically, for each class \(c \in \mathcal{C}_t\), we estimate its prototype \(\mu_c\) and covariance matrix \(\Sigma_c\), and sample pseudo-features \(\tilde{u}_c^i \sim \mathcal{N}(\mu_c,\Sigma_c)\) for \(i=1,\ldots,s_c\). The classifier is then optimized on these synthetic features via
\begin{equation}
\mathcal{L}_{\mathrm{CA}}
=
-\frac{1}{\sum_{c \in \mathcal{C}_t} s_c}
\sum_{c \in \mathcal{C}_t}
\sum_{i=1}^{s_c}
\log
\frac{\exp\!\left((\tilde{u}_c^i)^{\top} w_c\right)}
{\sum_{k \in \mathcal{C}_t} \exp\!\left((\tilde{u}_c^i)^{\top} w_k\right)}.
\end{equation}

\subsection{Theoretical Analysis}

We study the role of inter-layer relation drift in catastrophic forgetting in CIL. The forgetting after learning task \(t\) is defined as
\(
\mathcal{F}_t=\frac{1}{t-1}\sum_{s=1}^{t-1}\left(\mathcal{E}_{s}(W_t)-\mathcal{E}_{s}(W_s)\right),
\)
where \(\mathcal{E}_{s}(W)\) denotes the expected classification error on task \(s\) under model \(W\)~\cite{theroy_1:conf/iclr/00080DLS25,theory_2:conf/icml/LinJLS23}. This captures the performance degradation on previously learned tasks after updating the model with task \(t\).

\begin{lemma}
Let \(\gamma_s(x_s^i;W_t)\) denote the classification margin of sample \(x_s^i\) from task \(s\) under model \(W_t\), defined as
\(
\gamma_s(x_s^i;W_t)
=
(z_{s,t}^{L,i})^\top w_{y_s^i}
-
\max_{k\neq y_s^i}(z_{s,t}^{L,i})^\top w_k.
\)
Then
\[
\mathcal{E}_s(W_t)-\mathcal{E}_s(W_s)
\le
\frac{1}{\gamma_{\min,s}}
\;
\mathbb{E}_{x_s^i\sim\mathcal{D}_s}
\big[
\gamma_s(x_s^i;W_s)-\gamma_s(x_s^i;W_t)
\big],
\]
where \(\gamma_{\min,s}\) denotes the minimum margin over correctly classified samples from task \(s\).
\label{lemma_margin}
\end{lemma}

Lemma~\ref{lemma_margin} establishes that forgetting on a previous task is governed by the degradation of classification margins on its samples. Specifically, the increase in classification error is upper bounded by the expected reduction of margins under the updated model after incorporating subsequent tasks. This result indicates that the effect of subsequent task learning on previous tasks is manifested through changes in the decision boundaries of previously learned samples.

\begin{lemma}
For any sample \(x_s^i\) from task \(s\), define \(\Delta h_{s,t}^{\ell,i}=h_{s,t}^{\ell,i}-h_{s,s}^{\ell,i}\). Then
\(
\gamma_s(x_s^i;W_s)-\gamma_s(x_s^i;W_t)
\le
\|w_{y_s^i}-w_{k^*}\|
\left\|
\sum_{\ell=1}^{L}\Delta h_{s,t}^{\ell,i}
\right\|,
\)
where \(k^*=\arg\max_{k\neq y_s^i}(z_{s,t}^{L,i})^\top w_k\).
\label{lemma_residual}
\end{lemma}

Lemma~\ref{lemma_residual} shows that margin degradation is governed by the aggregated deviation of residual representations across layers. Specifically, the change in classification margins is determined by the cumulative effect of layer-wise residual outputs, where deviations introduced at different layers are aggregated through the forward propagation. As a result, even moderate variations at individual layers can accumulate and lead to a substantial shift in the final representation, thereby affecting the classification margin under the updated model. Aggregated residual deviation admits the following decomposition:
\begin{equation}
\left\|\sum_{\ell=1}^{L}\Delta h_{s,t}^{\ell,i}\right\|^2
=
\sum_{\ell=1}^{L}\|\Delta h_{s,t}^{\ell,i}\|^2
+
2\sum_{\ell<m}\langle \Delta h_{s,t}^{\ell,i}, \Delta h_{s,t}^{m,i}\rangle.
\label{eq:residual_expand}
\end{equation}
The first term measures the magnitude of layer-wise residual deviations, while the second term reflects the dependence among residual representations across layers. Even when deviations at individual layers are moderate, such dependence can accumulate and lead to a substantial change in the final representation. Accordingly, Theorem~\ref{theorem_residual} characterizes forgetting in terms of the aggregated deviation of residual representations.

\begin{theorem}
\label{theorem_residual}
The forgetting after learning task \(t\) satisfies
\begin{equation}
\begin{aligned}
\mathcal{F}_t
\le\;
&\frac{\|W_t^c\|_2}{t-1}
\sum_{s=1}^{t-1}
\frac{1}{\gamma_{\min,s}}
\;
\mathbb{E}_{x_s^i\sim\mathcal{D}_s}
\Bigg[
\Bigg(
\sum_{\ell=1}^{L}\|\Delta h_{s,t}^{\ell,i}\|^2 \\
&\;+\;
2\sum_{\ell<m}\langle \Delta h_{s,t}^{\ell,i}, \Delta h_{s,t}^{m,i}\rangle
\Bigg)^{1/2}
\Bigg]
\end{aligned}
\end{equation}
where \(\Delta h_{s,t}^{\ell,i}=h_{s,t}^{\ell,i}-h_{s,s}^{\ell,i}\) denotes the residual deviation at layer \(\ell\) for sample \(x_s^i\), \(\gamma_{\min,s}\) denotes the minimum classification margin over correctly classified samples from task \(s\), and \(W_t^c\) denotes the classifier after learning task \(t\).
\end{theorem}

\begin{definition}
For a sample \(x_s^i\), the inter-layer relation matrix \(R_{s,t}^i\in\mathbb{R}^{L\times L}\) is defined entry-wise as
\(
(R_{s,t}^i)_{\ell m}=\langle z_{s,t}^{\ell,i}, z_{s,t}^{m,i}\rangle,
\)
where \(\ell,m\in\{1,\ldots,L\}\).
\label{def_relation}
\end{definition}

The matrix \(R_{s,t}^i\) denotes the inter-layer relations among layer-wise representations under model \(W_t\). Since each \(z_{s,t}^{\ell,i}\) accumulates the residual outputs up to layer \(\ell\), the discrepancy between \(R_{s,t}^i\) and \(R_{s,s}^i\) quantifies how the inter-layer relations of sample \(x_s^i\) are altered after subsequent tasks are learned. Inter-layer relation drift controls the second term in Eq.~\eqref{eq:residual_expand}, leading to the bound
\begin{lemma}
For any sample \(x_s^i\),
\[
\sum_{\ell<m}
\left|
\langle \Delta h_{s,t}^{\ell,i}, \Delta h_{s,t}^{m,i}\rangle
\right|
\le
\frac{8(L-1)}{\lambda_{\min}(R_{s,s}^i)}
\;
\|R_{s,t}^i-R_{s,s}^i\|_F^2,
\]
where \(\lambda_{\min}(\cdot)>0\) denotes the smallest eigenvalue.
\label{lemma_relation}
\end{lemma}

Lemma~\ref{lemma_relation} shows that the component of aggregated residual deviation associated with inter-layer relations is upper bounded by inter-layer relation drift. This connects the residual-level characterization in Theorem~\ref{theorem_residual} to a structural quantity defined on layer-wise representations.

\begin{theorem}
\label{theorem_forgetting_relation}
The forgetting after learning task \(t\) satisfies
\begin{equation}
\begin{aligned}
\mathcal{F}_t
\le\;
&\frac{\|W_t^c\|_2}{t-1}
\sum_{s=1}^{t-1}
\frac{1}{\gamma_{\min,s}}
\cdot
\mathbb{E}_{x_s^i\sim\mathcal{D}_s}
\Bigg[
\Bigg(
\sum_{\ell=1}^{L}\|\Delta h_{s,t}^{\ell,i}\|^2 \\
&\;+\;
\frac{16(L-1)}{\lambda_{\min}(R_{s,s}^i)}
\|R_{s,t}^i-R_{s,s}^i\|_F^2
\Bigg)^{1/2}
\Bigg]
\end{aligned}
\end{equation}
where \(\Delta h_{s,t}^{\ell,i}=h_{s,t}^{\ell,i}-h_{s,s}^{\ell,i}\), \(\gamma_{\min,s}\) denotes the minimum classification margin on task \(s\), and \(W_t^c\) denotes the classifier after learning task \(t\).
\end{theorem}

Theorem~\ref{theorem_forgetting_relation} establishes a direct connection between inter-layer relation drift and forgetting. The bound shows that forgetting depends not only on the magnitude of layer-wise residual deviations, but also on the extent to which inter-layer relations are disrupted. A larger discrepancy between \(R_{s,t}^i\) and \(R_{s,s}^i\) leads to a larger upper bound on the performance degradation of task \(s\). 

\begin{table}[t]
\centering
\caption{Description of Symbols}
\label{tab:notation}
\resizebox{\linewidth}{!}{
\begin{tabular}{ll}
\toprule
\textbf{Sym.} & \textbf{Definition} \\
\midrule
\(t, s\) & task indices \\
\(T\) & total number of tasks \\
\(\mathcal{D}_t = \{(x_t^i, y_t^i)\}_{i=1}^{n_t}\) & dataset of task \(t\) \\
\(\mathcal{Y}_t\) & label set of task \(t\) \\
\(\mathcal{C}_t = \bigcup_{\tau=1}^{t} \mathcal{Y}_\tau\) & cumulative label space up to task \(t\) \\

\(W_0^\ell\) & pre-trained parameters at layer \(\ell\) \\
\(W_t^\ell\) & model parameters at layer \(\ell\) after task \(t\) \\
\(W_t\) & model after task \(t\) \\
\(W_t^c\) & classifier after task \(t\) \\
\(B_t^\ell A_t^\ell\) & low-rank adaptation at layer \(\ell\) for task \(t\) \\

\(z_{s,t}^{\ell,i}\) & representation of \(x_s^i\) at layer \(\ell\) under \(W_t\) \\
\(h_{s,t}^{\ell,i}\) & residual output at layer \(\ell\) \\
\(\Delta h_{s,t}^{\ell,i}\) & residual deviation between \(W_s\) and \(W_t\) \\

\(\gamma_s(x_s^i;W_t)\) & classification margin under \(W_t\) \\
\(\gamma_{\min,s}\) & minimum margin on task \(s\) \\
\(\mathcal{E}_s(W)\) & expected classification error on task \(s\) \\
\(\mathcal{F}_t\) & forgetting after learning task \(t\) \\

\(R_{s,t}^i\) & inter-layer relation matrix under \(W_t\) \\
\(\lambda_{\min}(\cdot)\) & minimum eigenvalue \\
\bottomrule
\end{tabular}
}
\end{table}

\subsection{Self-Rectifying inter-layer Relations LoRA}

Motivated by the theoretical analysis, we propose \textit{Self-Rectifying Inter-layer Relation LoRA} (SR$^2$-LoRA), which mitigates catastrophic forgetting by explicitly constraining inter-layer relation drift between consecutive models.

\textbf{Inter-layer Relation Construction.} For a sample \(x_t^i\) from the current task, we construct the inter-layer relation matrix from the layer-wise representations induced by the current model \(W_t\). Following Definition~\ref{def_relation}, the resulting matrix is denoted by \(R_{t,t}^i\in\mathbb{R}^{L\times L}\), with entries
\(
(R_{t,t}^i)_{\ell m}=\phi(z_{t,t}^{\ell,i},z_{t,t}^{m,i}),
\)
where \(\ell,m\in\{1,\ldots,L\}\) and \(\phi(\cdot,\cdot)\) denotes a feature similarity function. For the same sample \(x_t^i\), the previous model \(W_{t-1}\) analogously induces the relation matrix \(R_{t,t-1}^i\). The discrepancy between \(R_{t,t}^i\) and \(R_{t,t-1}^i\) therefore quantifies the inter-layer relation drift introduced by the task-\(t\) update on current-task samples.

\textbf{Relation Alignment.} A natural strategy for constraining inter-layer relation drift is to align \(R_{t,t}^i\) and \(R_{t,t-1}^i\) entry-wise. However, since the previous model \(W_{t-1}\) has not been optimized on current-task samples, the induced relation matrix \(R_{t,t-1}^i\) is subject to estimation perturbations. Direct entry-wise alignment is therefore sensitive to such perturbations and may enforce sample-specific deviations.

\begin{proposition}\label{prop:sv_stability}
Let \(R,\tilde{R}\in\mathbb{R}^{L\times L}\) satisfy \(\tilde{R}=R+E\), where \(E\) is a perturbation matrix. Let \(\operatorname{sv}_i(\cdot)\) denote the \(i\)-th singular value. Then, for each \(i\in\{1,\dots,L\}\),
\[
\left| \operatorname{sv}_i(\tilde{R})-\operatorname{sv}_i(R) \right| \le \|E\|_F,
\]
where \(\|\cdot\|_F\) denotes the Frobenius norm.
\end{proposition}

Proposition~\ref{prop:sv_stability} indicates that singular values remain stable under bounded perturbations, in contrast to entry-wise quantities which can vary arbitrarily. This property makes singular values a more robust characterization of inter-layer relations under estimation noise. Accordingly, we align the singular values of \(R_{t,t-1}^i\) and \(R_{t,t}^i\). Let
\(
R_{t,t-1}^i=U_{t,t-1}^i\Lambda_{t,t-1}^i(V_{t,t-1}^i)^\top
\)
and
\(
R_{t,t}^i=U_{t,t}^i\Lambda_{t,t}^i(V_{t,t}^i)^\top
\)
denote their singular value decompositions. The relation alignment loss for sample \(x_t^i\) is
\begin{equation}
\begin{aligned}
\mathcal{L}_{\mathrm{align}}^{i}
&=
\frac{1}{L}
\sum_{\ell=1}^{L}
\rho\!\left(
\operatorname{sv}_{\ell}(R_{t,t-1}^{i})
-
\operatorname{sv}_{\ell}(R_{t,t}^{i})
\right), \\
\rho(\delta)
&=
\begin{cases}
\frac{1}{2}\delta^2, & |\delta|<1,\\
|\delta|-\frac{1}{2}, & \text{otherwise}.
\end{cases}
\end{aligned}
\label{eq:align}
\end{equation}
where \(\operatorname{sv}_{\ell}(R)\) denotes the \(\ell\)-th singular value of \(R\). Since inter-layer relations are defined for each sample, the alignment is performed in a sample-wise manner. The overall objective at task \(t\) combines the standard classification loss with the relation alignment loss:
\begin{equation}
\mathcal{L}_{\mathrm{total}}
=
\mathcal{L}_{\mathrm{CE}}
+
\lambda \frac{1}{n_t}\sum_{i=1}^{n_t}\mathcal{L}_{\mathrm{align}}^{i},
\label{eq:sr2}
\end{equation}
where \(\lambda\) is a trade-off coefficient that balances classification learning and inter-layer relation alignment. The overall optimization procedure is summarized in Algorithm~\ref{alg:sr2_lora}.

\begin{algorithm}[t]
\caption{SR$^2$-LoRA Training at Task \(t\)}
\label{alg:sr2_lora}
\begin{algorithmic}[1]
\STATE \textbf{Input:} Dataset \(\mathcal{D}_t\), previous model \(W_{t-1}\), current model \(W_t\), trade-off coefficient \(\lambda\), learning rate \(\eta\), maximum iterations \(I_{\max}\)
\STATE \textbf{Output:} Updated model \(W_t\)

\STATE Initialize the trainable LoRA parameters in \(W_t\)

\FOR{\(iter = 1\) to \(I_{\max}\)}
    \STATE Sample a mini-batch \(\mathcal{B}_t \subset \mathcal{D}_t\)
    \STATE Compute the classification loss \(\mathcal{L}_{\mathrm{CE}}\) under \(W_t\)

    \FOR{each sample \(x_t^i \in \mathcal{B}_t\)}
        \STATE Perform forward passes with \(W_{t-1}\) and \(W_t\) to obtain \(\{z_{t,t-1}^{\ell,i}\}_{\ell=1}^{L}\) and \(\{z_{t,t}^{\ell,i}\}_{\ell=1}^{L}\)
        \STATE Construct \(R_{t,t-1}^{i}\) and \(R_{t,t}^{i}\) according to Definition~\ref{def_relation}
        \STATE Compute the singular values of \(R_{t,t-1}^{i}\) and \(R_{t,t}^{i}\)
        \STATE Compute \(\mathcal{L}_{\mathrm{align}}^{i}\) according to Eq.~(\ref{eq:align})
        \STATE Accumulate \(\mathcal{L}_{\mathrm{align}} \leftarrow \mathcal{L}_{\mathrm{align}} + \mathcal{L}_{\mathrm{align}}^{i}\)
    \ENDFOR

    \STATE Compute \(\mathcal{L}_{\mathrm{total}} \leftarrow \mathcal{L}_{\mathrm{CE}} + \lambda |\mathcal{B}_t|^{-1}\mathcal{L}_{\mathrm{align}}\)
    \STATE Update the trainable parameters in \(W_t\) via gradient descent with learning rate \(\eta\)
\ENDFOR

\STATE \textbf{return} \(W_t\)
\end{algorithmic}
\end{algorithm}

\section{Experiments}

\subsection{Experimental Setup}

\textbf{Datasets.} Following standard CIL evaluation settings~\cite{liu2025lora}, we evaluate the proposed method on four widely used benchmarks, including CIFAR-100~\cite{krizhevsky2009learning}, ImageNet-R~\cite{hendrycks2021many}, CUB-200~\cite{wah2011caltech}, and ImageNet-A~\cite{hendrycks2021natural}. CIFAR-100 consists of 100 categories, whereas ImageNet-R, CUB-200, and ImageNet-A each contain 200 categories. All datasets are evaluated under 5-, 10-, 20-, and 50-task settings.

\textbf{Evaluation Metrics.} Following CIL protocols~\cite{zhou2024expandable, liu2025lora}, performance is assessed using the final accuracy \(\mathcal{A}_{Last}\) and the average accuracy \(\mathcal{A}_{Avg}\). Let \(a_{i,j}\) denote the test accuracy on the \(j\)-th task \((j \le i)\) after learning the \(i\)-th task. The average accuracy after task \(i\) is defined as \(\mathcal{A}_i = \frac{1}{i}\sum_{j=1}^{i} a_{i,j}\). The final accuracy is defined as \(\mathcal{A}_{Last} = \frac{1}{T}\sum_{j=1}^{T} a_{T,j}\), where \(T\) denotes the total number of tasks. The overall average accuracy is given by \(\mathcal{A}_{Avg} = \frac{1}{T}\sum_{i=1}^{T} \mathcal{A}_i\).

\textbf{Baselines.} We compare with representative PTM-based CIL methods from five categories, including selection-based methods such as CODA-Prompt~\cite{smith2023coda}, MOS~\cite{sun2025mos}, and CL-LoRA~\cite{he2025cl}, prototype-based methods such as SSIAT~\cite{tan2024semantically}, EASE~\cite{zhou2024expandable}, and LoRA-DRS~\cite{liu2025lora}, covariance-based methods such as MACIL~\cite{wu2025navigating}, orthogonality-based methods such as InfLoRA~\cite{liang2024inflora}, and fine-tuning methods such as SLCA~\cite{zhang2023slca}.

\textbf{Implementation Details.}
Following prior work~\cite{liang2024inflora, liu2025lora}, all experiments are conducted using a single NVIDIA RTX 4090 GPU. We adopt ViT-B/16~\cite{dosovitskiy2020image} with 12 transformer blocks, pretrained on ImageNet-21K~\cite{russakovsky2015imagenet}, as the backbone architecture. LoRA modules with rank \(r=10\) are integrated into the key and value projections of all attention layers. The model is optimized using Adam with a learning rate of \(5\times10^{-4}\) and a cosine annealing schedule, with a batch size of 128. Each task is trained for 50 epochs on ImageNet-R and ImageNet-A, and 20 epochs on CIFAR-100 and CUB-200. Following standard evaluation protocols, all results are reported as the mean and standard deviation over three runs with different random seeds.

\begin{table*}[t]
\centering
\caption{Performance on different datasets under varying task-length settings. All values are reported as percentages, and the best results are highlighted in bold. All methods use the same ViT-B/16-IN21K backbone, random seeds, and class orders.}
\label{tab:main_results}
\resizebox{\textwidth}{!}{
\begin{tabular}{llcccccccccc}
\toprule
\multirow{2}{*}{Dataset} & \multirow{2}{*}{Method} & \multicolumn{2}{c}{5 Tasks} & \multicolumn{2}{c}{10 Tasks} & \multicolumn{2}{c}{20 Tasks} & \multicolumn{2}{c}{50 Tasks} \\ \cmidrule(lr){3-4} \cmidrule(lr){5-6} \cmidrule(lr){7-8} \cmidrule(lr){9-10}
 & & $\mathcal{A}_{Avg}$ & $\mathcal{A}_{Last}$ & $\mathcal{A}_{Avg}$ & $\mathcal{A}_{Last}$ & $\mathcal{A}_{Avg}$ & $\mathcal{A}_{Last}$ & $\mathcal{A}_{Avg}$ & $\mathcal{A}_{Last}$\\ \midrule
\multirow{2}{*}{CIFAR100} & Vanilla & 90.92\tiny$\pm$0.41 & 84.21\tiny$\pm$1.19 & 88.26\tiny$\pm$0.92 & 79.94\tiny$\pm$1.29 & 84.50\tiny$\pm$0.27 & 73.23\tiny$\pm$0.34 & 76.39\tiny$\pm$4.08 & 59.73\tiny$\pm$4.05 \\
 & Ours    & \textbf{94.91\tiny$\pm$0.34} & \textbf{92.26\tiny$\pm$0.26} & \textbf{94.60\tiny$\pm$0.08} & \textbf{91.98\tiny$\pm$0.10} & \textbf{94.08\tiny$\pm$0.12} & \textbf{90.45\tiny$\pm$0.13} & \textbf{93.21\tiny$\pm$0.43} & \textbf{89.12\tiny$\pm$0.74}  \\ \midrule
\multirow{2}{*}{CUB-200}  & Vanilla  & 85.80\tiny$\pm$0.38 & 75.47\tiny$\pm$0.13 & 79.48\tiny$\pm$1.21 & 67.40\tiny$\pm$2.12 & 70.67\tiny$\pm$1.28 & 47.06\tiny$\pm$4.90 & 60.95\tiny$\pm$3.75 & 38.08\tiny$\pm$3.90 \\
 & Ours     & \textbf{93.22\tiny$\pm$0.32} & \textbf{90.09\tiny$\pm$0.34} & \textbf{94.01\tiny$\pm$0.79} & \textbf{90.37\tiny$\pm$0.12} & \textbf{94.32\tiny$\pm$0.11} & \textbf{89.71\tiny$\pm$0.21} & \textbf{94.30\tiny$\pm$0.88} & \textbf{89.44\tiny$\pm$0.29} \\ \midrule
\multirow{2}{*}{ImageNet-R} & Vanilla  & 80.72\tiny$\pm$0.34 & 72.78\tiny$\pm$0.67 & 76.52\tiny$\pm$0.52 & 64.66\tiny$\pm$0.89 & 69.76\tiny$\pm$0.11 & 56.71\tiny$\pm$1.45 & 61.24\tiny$\pm$1.27 & 44.40\tiny$\pm$0.77\\
 & Ours     & \textbf{86.01\tiny$\pm$0.39} & \textbf{81.72\tiny$\pm$0.27} & \textbf{85.93\tiny$\pm$0.54} & \textbf{81.14\tiny$\pm$0.23} & \textbf{84.32\tiny$\pm$0.53} & \textbf{79.54\tiny$\pm$0.45} & \textbf{80.51\tiny$\pm$0.70} & \textbf{74.85\tiny$\pm$0.27}\\ \midrule
\multirow{2}{*}{ImageNet-A} & Vanilla  & 59.11\tiny$\pm$0.15 & 46.06\tiny$\pm$3.25 & 52.35\tiny$\pm$1.31 & 36.56\tiny$\pm$2.66 & 44.76\tiny$\pm$0.78 & 26.13\tiny$\pm$1.40 & 36.90\tiny$\pm$1.76 & 19.93\tiny$\pm$2.29\\
 & Ours    & \textbf{70.67\tiny$\pm$1.61} & \textbf{63.27\tiny$\pm$0.93} & \textbf{70.03\tiny$\pm$1.02} & \textbf{63.47\tiny$\pm$0.78} & \textbf{69.41\tiny$\pm$0.12} & \textbf{61.01\tiny$\pm$0.79} & \textbf{66.47\tiny$\pm$1.23} & \textbf{57.08\tiny$\pm$1.25} \\ \bottomrule
\end{tabular}
}
\end{table*}

\begin{figure*}[t]
  \begin{center}
    \centerline{\includegraphics[width=\textwidth]{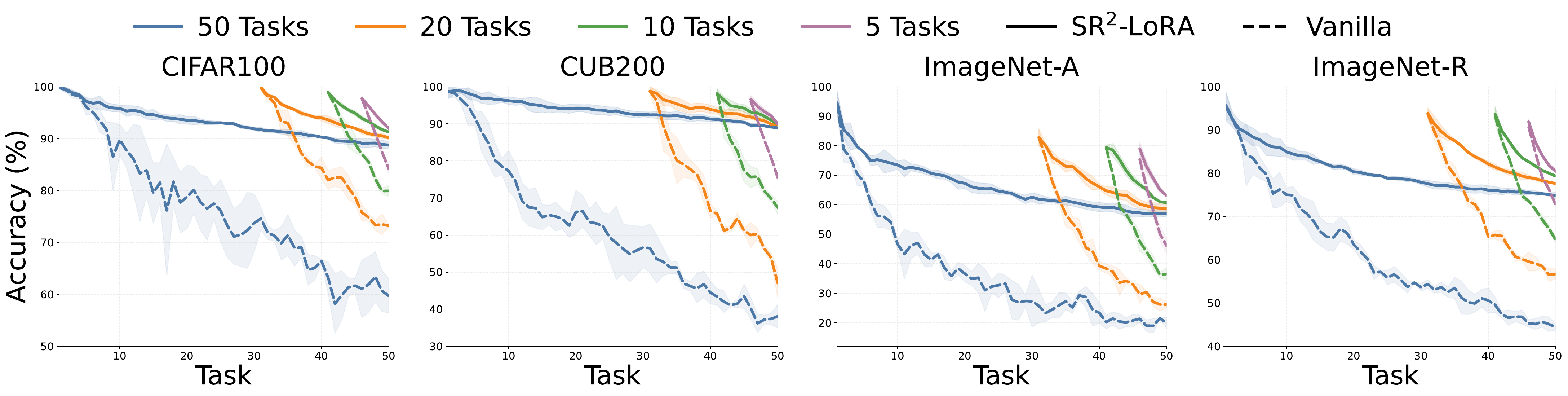}}
    \caption{Performance after each task on all seen tasks across datasets with varying task lengths. All methods use the same ViT-B/16-IN21K backbone, random seeds, and class orders.}
    \vspace{-10pt}
    \label{fig:main_acc}
  \end{center}
\end{figure*}

\subsection{Main Results}

We evaluate SR$^2$-LoRA on multiple CIL benchmarks. Table~\ref{tab:main_results} reports the comparison between SR$^2$-LoRA and vanilla training under different task lengths. Table~\ref{tab:four_benchmarks_results} presents comparisons with representative methods under the 10-task setting, while Tables~\ref{tab:cifar_cub} and~\ref{tab:iamgenet} further report the results under the 20- and 50-task settings, respectively. Across nearly all settings, SR$^2$-LoRA consistently improves model performance over vanilla training, and the improvement becomes increasingly pronounced as the number of tasks grows. This result indicates that constraining inter-layer relation drift effectively mitigates catastrophic forgetting, especially in long task sequences.

\begin{table}[t]
\centering
\caption{Average performance across different task-length settings on multiple datasets. All values are reported as percentages. All methods use the same ViT-B/16-IN21K backbone, random seeds, and class orders.}
\label{tab:avg_only}
\resizebox{\linewidth}{!}{
\begin{tabular}{llcc}
\toprule
Dataset & Method 
& $\mathcal{A}_{Avg}$ 
& $\mathcal{A}_{Last}$ \\
\midrule
\multirow{2}{*}{CIFAR100} 
& Vanilla & 85.02$\pm$6.33 & 74.28$\pm$10.70 \\
& Ours    & 94.20$\pm$0.74 & 90.95$\pm$1.45 \\
\midrule
\multirow{2}{*}{CUB-200}
& Vanilla & 74.23$\pm$10.81 & 57.00$\pm$17.38 \\
& Ours & 93.96$\pm$0.51 & 89.90$\pm$0.41 \\
\midrule
\multirow{2}{*}{ImageNet-R}
& Vanilla & 72.06$\pm$8.51 & 59.64$\pm$12.09 \\
& Ours & 84.19$\pm$2.57 & 79.31$\pm$3.11 \\
\midrule
\multirow{2}{*}{ImageNet-A}
& Vanilla & 48.28$\pm$9.59 & 32.17$\pm$11.53 \\
& Ours & 69.14$\pm$1.85 & 61.20$\pm$2.96 \\
\bottomrule
\end{tabular}
}
\end{table}

\begin{table*}[t]
\caption{Last and average performance on CIFAR-100 and CUB-200 under long task sequences with 20 and 50 tasks. Results are reported as percentages, and the best values are highlighted in bold. All methods use the same ViT-B/16-IN21K backbone, random seeds, and class orders.}
\label{tab:cifar_cub}
\centering
\resizebox{\textwidth}{!}{
\begin{tabular}{@{}lcccccccc@{}}
\toprule
\multirow{3}{*}{\textbf{Method}} & \multicolumn{4}{c}{\textbf{CIFAR-100}} & \multicolumn{4}{c}{\textbf{CUB-200}} \\
\cmidrule(lr){2-5} \cmidrule(lr){6-9}
& \multicolumn{2}{c}{20} & \multicolumn{2}{c}{50} & \multicolumn{2}{c}{20} & \multicolumn{2}{c}{50} \\
& \multicolumn{1}{c}{$\mathcal{A}_{Last}$} & \multicolumn{1}{c}{$\mathcal{A}_{Avg}$} & \multicolumn{1}{c}{$\mathcal{A}_{Last}$} & \multicolumn{1}{c}{$\mathcal{A}_{Avg}$} & \multicolumn{1}{c}{$\mathcal{A}_{Last}$} & \multicolumn{1}{c}{$\mathcal{A}_{Avg}$} & \multicolumn{1}{c}{$\mathcal{A}_{Last}$} & \multicolumn{1}{c}{$\mathcal{A}_{Avg}$} \\
\midrule
\multicolumn{1}{l}{CODA-Prompt~\cite{smith2023coda}} 
& \multicolumn{1}{c}{81.36\tiny$\pm$0.88} & \multicolumn{1}{c}{88.17\tiny$\pm$0.61} 
& \multicolumn{1}{c}{55.45\tiny$\pm$0.48} & \multicolumn{1}{c}{68.39\tiny$\pm$0.53} 
& 66.41\tiny$\pm$0.81 & 78.10\tiny$\pm$1.87 
& 46.25\tiny$\pm$0.68 & 63.25\tiny$\pm$2.69 \\
\multicolumn{1}{l}{SLCA~\cite{zhang2023slca}} 
& \multicolumn{1}{c}{89.62\tiny$\pm$0.18} & \multicolumn{1}{c}{93.03\tiny$\pm$1.09} 
& \multicolumn{1}{c}{87.90\tiny$\pm$0.17} & \multicolumn{1}{c}{91.96\tiny$\pm$1.30} 
& 82.48\tiny$\pm$0.53 & 90.14\tiny$\pm$1.02 
& 78.47\tiny$\pm$1.85 & 88.38\tiny$\pm$0.24 \\
\multicolumn{1}{l}{InfLoRA~\cite{liang2024inflora}} 
& \multicolumn{1}{c}{81.59\tiny$\pm$0.94} & \multicolumn{1}{c}{87.33\tiny$\pm$2.32} 
& \multicolumn{1}{c}{55.19\tiny$\pm$2.83} & \multicolumn{1}{c}{69.96\tiny$\pm$3.55} 
& - & - & - & - \\
\multicolumn{1}{l}{EASE~\cite{zhou2024expandable}} 
& \multicolumn{1}{c}{86.32\tiny$\pm$0.48} & \multicolumn{1}{c}{90.83\tiny$\pm$1.14} 
& \multicolumn{1}{c}{77.43\tiny$\pm$3.11} & \multicolumn{1}{c}{84.59\tiny$\pm$1.22} 
& 84.51\tiny$\pm$1.67 & 91.02\tiny$\pm$1.02 
& 86.46\tiny$\pm$0.14 & 92.38\tiny$\pm$0.25 \\
\multicolumn{1}{l}{SSIAT~\cite{tan2024semantically}} 
& \multicolumn{1}{c}{90.07\tiny$\pm$0.56} & \multicolumn{1}{c}{93.54\tiny$\pm$0.83} 
& \multicolumn{1}{c}{\underline{87.33\tiny$\pm$1.46}} & \multicolumn{1}{c}{\underline{91.63\tiny$\pm$0.89}} 
& 89.06\tiny$\pm$0.66 & 93.52\tiny$\pm$0.55 
& 83.64\tiny$\pm$1.67 & 91.84\tiny$\pm$0.47 \\
\multicolumn{1}{l}{CL-LoRA~\cite{he2025cl}} 
& \multicolumn{1}{c}{83.75\tiny$\pm$1.39} & \multicolumn{1}{c}{88.98\tiny$\pm$1.87} 
& \multicolumn{1}{c}{71.14\tiny$\pm$2.44} & \multicolumn{1}{c}{78.80\tiny$\pm$2.72} 
& - & - & - & - \\
\multicolumn{1}{l}{LoRA-DRS~\cite{liu2025lora}} 
& \multicolumn{1}{c}{88.76\tiny$\pm$0.22} & \multicolumn{1}{c}{92.35\tiny$\pm$0.90} 
& \multicolumn{1}{c}{86.51\tiny$\pm$1.05} & \multicolumn{1}{c}{91.17\tiny$\pm$0.97} 
& 87.50\tiny$\pm$0.26 & 92.66\tiny$\pm$0.57 
& 86.97\tiny$\pm$0.13 & 92.70\tiny$\pm$0.30 \\
\multicolumn{1}{l}{MOS~\cite{sun2025mos}} 
& \multicolumn{1}{c}{89.48\tiny$\pm$0.39} & \multicolumn{1}{c}{92.97\tiny$\pm$1.01} 
& \multicolumn{1}{c}{87.05\tiny$\pm$0.90} & \multicolumn{1}{c}{91.44\tiny$\pm$1.21} 
& \underline{89.42\tiny$\pm$0.22} & \underline{93.66\tiny$\pm$0.42} 
& \underline{89.28\tiny$\pm$0.25} & \underline{93.47\tiny$\pm$0.43} \\
\multicolumn{1}{l}{MACIL~\cite{wu2025navigating}} 
& \multicolumn{1}{c}{\underline{90.31\tiny$\pm$0.17}} & \multicolumn{1}{c}{\underline{93.47\tiny$\pm$0.78}} 
& \multicolumn{1}{c}{85.09\tiny$\pm$0.87} & \multicolumn{1}{c}{90.89\tiny$\pm$1.24} 
& 88.63\tiny$\pm$0.61 & 93.52\tiny$\pm$0.43 
& 82.06\tiny$\pm$0.48 & 91.04\tiny$\pm$0.41 \\
\midrule
\multicolumn{1}{l}{SR$^2$-LoRA} & \textbf{90.45\tiny$\pm$0.13}  & \textbf{94.08\tiny$\pm$0.12} & \textbf{89.12\tiny$\pm$0.74} & \textbf{93.21\tiny$\pm$0.43} & \textbf{89.71\tiny$\pm$0.21} & \textbf{94.32\tiny$\pm$0.11} & \textbf{89.44\tiny$\pm$0.29} & \textbf{94.30\tiny$\pm$0.88}\\
\bottomrule
\end{tabular}}
\end{table*}

\begin{table*}[h]
\caption{Last and average performance on ImageNet-R and ImageNet-A under long task sequences with 20 and 50 tasks. Results are reported as percentages, and the best values are highlighted in bold. All methods use the same ViT-B/16-IN21K backbone, random seeds, and class orders.} 
\label{tab:iamgenet}
\centering
\resizebox{\textwidth}{!}{
\begin{tabular}{@{}lcccccccc@{}}
\toprule
\multirow{3}{*}{\textbf{Method}} & \multicolumn{4}{c}{\textbf{ImageNet-R}} & \multicolumn{4}{c}{\textbf{ImageNet-A}} \\
\cmidrule(lr){2-5} \cmidrule(lr){6-9} & \multicolumn{2}{c}{20} & \multicolumn{2}{c}{50} & \multicolumn{2}{c}{20} & \multicolumn{2}{c}{50} \\
& \multicolumn{1}{c}{$\mathcal{A}_{Last}$} & \multicolumn{1}{c}{$\mathcal{A}_{Avg}$} & \multicolumn{1}{c}{$\mathcal{A}_{Last}$} & \multicolumn{1}{c}{$\mathcal{A}_{Avg}$} & \multicolumn{1}{c}{$\mathcal{A}_{Last}$} & \multicolumn{1}{c}{$\mathcal{A}_{Avg}$} & \multicolumn{1}{c}{$\mathcal{A}_{Last}$} & \multicolumn{1}{c}{$\mathcal{A}_{Avg}$} \\
\midrule
\multicolumn{1}{l}{CODA-Prompt~\cite{smith2023coda}} 
& \multicolumn{1}{c}{69.96\tiny$\pm$0.50} & \multicolumn{1}{c}{75.34\tiny$\pm$0.85} & \multicolumn{1}{c}{48.89\tiny$\pm$0.90} & \multicolumn{1}{c}{55.59\tiny$\pm$2.67} & 45.40\tiny$\pm$1.39 & 54.55\tiny$\pm$0.86 & 30.35\tiny$\pm$0.96 & 41.53\tiny$\pm$0.64 \\
\multicolumn{1}{l}{SLCA~\cite{zhang2023slca}} 
& \multicolumn{1}{c}{75.53\tiny$\pm$0.42} & \multicolumn{1}{c}{80.65\tiny$\pm$1.16} & \multicolumn{1}{c}{68.95\tiny$\pm$4.46} & \multicolumn{1}{c}{71.15\tiny$\pm$10.57} & 55.01\tiny$\pm$2.66 & 63.59\tiny$\pm$2.20 & 49.31\tiny$\pm$0.93 & 56.72\tiny$\pm$1.82 \\
\multicolumn{1}{l}{InfLoRA~\cite{liang2024inflora}} 
& \multicolumn{1}{c}{73.01\tiny$\pm$0.82} & \multicolumn{1}{c}{79.76\tiny$\pm$0.83} & \multicolumn{1}{c}{61.91\tiny$\pm$1.31} & \multicolumn{1}{c}{71.20\tiny$\pm$0.54} 
& - & - & - & - \\
\multicolumn{1}{l}{EASE~\cite{zhou2024expandable}} 
& \multicolumn{1}{c}{73.78\tiny$\pm$0.47} & \multicolumn{1}{c}{80.28\tiny$\pm$0.67} & \multicolumn{1}{c}{68.53\tiny$\pm$0.04} & \multicolumn{1}{c}{75.49\tiny$\pm$1.06} & 50.32\tiny$\pm$2.11 & 61.77\tiny$\pm$1.60 & 37.06\tiny$\pm$0.50 & 49.80\tiny$\pm$0.24 \\
\multicolumn{1}{l}{SSIAT~\cite{tan2024semantically}} 
& \multicolumn{1}{c}{{78.31\tiny$\pm$0.53}} & \multicolumn{1}{c}{82.35\tiny$\pm$0.52} & \multicolumn{1}{c}{\underline{74.52\tiny$\pm$0.35}} & \multicolumn{1}{c}{\underline{79.01\tiny$\pm$0.54}} & \underline{60.52\tiny$\pm$2.07} & \underline{68.87\tiny$\pm$2.29} & 51.44\tiny$\pm$0.57 & 61.63\tiny$\pm$2.61 \\
\multicolumn{1}{l}{CL-LoRA~\cite{he2025cl}} 
& \multicolumn{1}{c}{77.20\tiny$\pm$0.66} & \multicolumn{1}{c}{{83.45\tiny$\pm$0.56}} & \multicolumn{1}{c}{68.36\tiny$\pm$0.82} & \multicolumn{1}{c}{76.80\tiny$\pm$1.27} 
& - & - & - & - \\
\multicolumn{1}{l}{LoRA-DRS~\cite{liu2025lora}} 
& \multicolumn{1}{c}{74.96\tiny$\pm$0.17} & \multicolumn{1}{c}{80.66\tiny$\pm$0.67} & \multicolumn{1}{c}{72.17\tiny$\pm$0.27} & \multicolumn{1}{c}{78.05\tiny$\pm$0.70} & 55.02\tiny$\pm$1.74 & 64.52\tiny$\pm$1.16 & \underline{53.28\tiny$\pm$1.75} & \underline{63.44\tiny$\pm$2.78} \\
\multicolumn{1}{l}{MOS~\cite{sun2025mos}} 
& \multicolumn{1}{c}{75.04\tiny$\pm$0.75} & \multicolumn{1}{c}{80.39\tiny$\pm$0.87} & \multicolumn{1}{c}{66.92\tiny$\pm$0.34} & \multicolumn{1}{c}{74.81\tiny$\pm$0.41} & 55.26\tiny$\pm$0.92 & 64.53\tiny$\pm$1.02 & 52.29\tiny$\pm$1.58 & 62.91\tiny$\pm$1.10 \\
\multicolumn{1}{l}{MACIL~\cite{wu2025navigating}} 
& \multicolumn{1}{c}{\underline{79.46\tiny$\pm$0.15}} & \multicolumn{1}{c}{\underline{84.25\tiny$\pm$0.51}} & \multicolumn{1}{c}{70.10\tiny$\pm$1.02} & \multicolumn{1}{c}{77.47\tiny$\pm$0.55} & {59.40\tiny$\pm$0.76} & 67.79\tiny$\pm$1.51 & 47.86\tiny$\pm$1.78 & 59.96\tiny$\pm$1.44 \\
\midrule
\multicolumn{1}{l}{SR$^2$-LoRA} & \textbf{79.54\tiny$\pm$0.45}  & \textbf{84.32\tiny$\pm$0.53} & \textbf{74.85\tiny$\pm$0.27} & \textbf{80.51\tiny$\pm$1.06} & \textbf{61.01\tiny$\pm$0.79} & \textbf{69.41\tiny$\pm$0.12} & \textbf{57.08\tiny$\pm$1.25} & \textbf{66.47\tiny$\pm$1.23}\\
\bottomrule
  \end{tabular}}
\end{table*}

The superiority of SR$^2$-LoRA is consistent with the theoretical analysis. As the number of tasks increases, inter-layer relation drift accumulates, which progressively reduces the classification margins of previous tasks and leads to more severe performance degradation. By explicitly constraining such drift, SR$^2$-LoRA better preserves the inter-layer relations established for previous tasks, thereby alleviating the disruption caused by subsequent task updates. This effect becomes more evident in longer task sequences, where the accumulation of inter-layer relation drift is more substantial. This observation is further supported by the training dynamics. Fig~\ref{fig:main_acc} shows the performance trend on previously seen tasks during training. Under different task lengths, SR$^2$-LoRA consistently outperforms vanilla training throughout the learning process. The advantage also becomes more pronounced as training proceeds over longer task sequences, further confirming the benefit of constraining inter-layer relation drift. We finally evaluate the stability of SR$^2$-LoRA under different task lengths. As shown in Table~\ref{tab:avg_only}, the proposed method maintains stable performance as the number of tasks increases, while the gap across different task-length settings remains small. This result suggests that SR$^2$-LoRA is robust to variations in task length and further supports its effectiveness in long task scenarios.

\begin{table*}[t]
 \caption{Last and average performance on four benchmark datasets under the 10-task setting. The mean and standard deviation over three runs are reported. All methods use the same ViT-B/16-IN21K backbone, random seeds, and class orders.}
  \centering
  \resizebox{\textwidth}{!}{
  \begin{tabular}{lcccccccc}
    \toprule
    \multirow{2}{*}{\textbf{Method}} & \multicolumn{2}{c}{\textbf{CIFAR-100}} & \multicolumn{2}{c}{\textbf{ImageNet-A}} & \multicolumn{2}{c}{\textbf{ImageNet-R}} & \multicolumn{2}{c}{\textbf{CUB-200}}\\
    & \multicolumn{1}{c}{$\mathcal{A}_{Last}$} & \multicolumn{1}{c}{$\mathcal{A}_{Avg}$} &  \multicolumn{1}{c}{$\mathcal{A}_{Last}$} & \multicolumn{1}{c}{$\mathcal{A}_{Avg}$} & \multicolumn{1}{c}{$\mathcal{A}_{Last}$} & \multicolumn{1}{c}{$\mathcal{A}_{Avg}$} & \multicolumn{1}{c}{$\mathcal{A}_{Last}$} & \multicolumn{1}{c}{$\mathcal{A}_{Avg}$} \\
    \midrule
    \multicolumn{1}{l}{L2P~\cite{wang2022learning}} & \multicolumn{1}{c}{84.06\tiny$\pm$0.88} & \multicolumn{1}{c}{88.26\tiny$\pm$1.34} &  \multicolumn{1}{c}{44.04\tiny$\pm$0.93} & \multicolumn{1}{c}{51.24\tiny$\pm$2.26} & \multicolumn{1}{c}{72.34\tiny$\pm$0.17} & \multicolumn{1}{c}{77.36\tiny$\pm$0.64} & \multicolumn{1}{c}{67.02\tiny$\pm$1.90} & \multicolumn{1}{c}{79.62\tiny$\pm$1.60}  \\
    \multicolumn{1}{l}{DualPrompt~\cite{wang2022dualprompt}} & \multicolumn{1}{c}{86.93\tiny$\pm$0.24} & \multicolumn{1}{c}{91.13\tiny$\pm$0.32} &  \multicolumn{1}{c}{53.19\tiny$\pm$0.74} & \multicolumn{1}{c}{64.59\tiny$\pm$0.08} & \multicolumn{1}{c}{69.10\tiny$\pm$0.62} & \multicolumn{1}{c}{74.28\tiny$\pm$0.66} & \multicolumn{1}{c}{68.48\tiny$\pm$0.47} & \multicolumn{1}{c}{80.59\tiny$\pm$1.50} \\
    \multicolumn{1}{l}{CODA-Prompt~\cite{smith2023coda}} & \multicolumn{1}{c}{83.21\tiny$\pm$3.39} & \multicolumn{1}{c}{87.71\tiny$\pm$3.17} &  \multicolumn{1}{c}{52.08\tiny$\pm$0.12} & \multicolumn{1}{c}{63.92\tiny$\pm$0.12} & \multicolumn{1}{c}{73.31\tiny$\pm$0.50} & \multicolumn{1}{c}{78.47\tiny$\pm$0.53} & \multicolumn{1}{c}{77.23\tiny$\pm$1.12} & \multicolumn{1}{c}{81.90\tiny$\pm$0.85} \\
    \multicolumn{1}{l}{SLCA~\cite{zhang2023slca}} & \multicolumn{1}{c}{91.26\tiny$\pm$0.37} & \multicolumn{1}{c}{94.29\tiny$\pm$0.92} &  \multicolumn{1}{c}{61.05\tiny$\pm$0.63} & \multicolumn{1}{c}{68.88\tiny$\pm$2.31} & \multicolumn{1}{c}{79.35\tiny$\pm$0.28} & \multicolumn{1}{c}{83.29\tiny$\pm$0.46} & \multicolumn{1}{c}{84.68\tiny$\pm$0.09} & \multicolumn{1}{c}{90.77\tiny$\pm$0.79} \\
    \multicolumn{1}{l}{InfLoRA~\cite{liang2024inflora}} & \multicolumn{1}{c}{86.20\tiny$\pm$0.70} & \multicolumn{1}{c}{90.58\tiny$\pm$1.52} &  \multicolumn{1}{c}{47.75\tiny$\pm$0.51} & \multicolumn{1}{c}{58.13\tiny$\pm$0.56} & \multicolumn{1}{c}{75.88\tiny$\pm$0.32} & \multicolumn{1}{c}{81.90\tiny$\pm$0.65} & \multicolumn{1}{c}{69.04\tiny$\pm$1.25} & \multicolumn{1}{c}{81.83\tiny$\pm$0.45} \\
    \multicolumn{1}{l}{EASE~\cite{zhou2024expandable}} & \multicolumn{1}{c}{88.22\tiny$\pm$0.44} & \multicolumn{1}{c}{92.02\tiny$\pm$0.76} &  \multicolumn{1}{c}{54.93\tiny$\pm$1.14} & \multicolumn{1}{c}{63.92\tiny$\pm$0.76} & \multicolumn{1}{c}{75.91\tiny$\pm$0.17} & \multicolumn{1}{c}{81.38\tiny$\pm$0.29} & \multicolumn{1}{c}{85.04\tiny$\pm$1.42} & \multicolumn{1}{c}{90.93\tiny$\pm$1.03} \\
    \multicolumn{1}{l}{SSIAT~\cite{tan2024semantically}} & \multicolumn{1}{c}{91.48\tiny$\pm$0.24} & \multicolumn{1}{c}{94.28\tiny$\pm$0.90} &  \multicolumn{1}{c}{62.58\tiny$\pm$1.58} & \multicolumn{1}{c}{\textbf{70.73\tiny$\pm$1.44}} & \multicolumn{1}{c}{79.54\tiny$\pm$0.24} & \multicolumn{1}{c}{83.67\tiny$\pm$0.57} & \multicolumn{1}{c}{89.83\tiny$\pm$0.53} & \multicolumn{1}{c}{93.76\tiny$\pm$0.52} \\
    \multicolumn{1}{l}{CL-LoRA~\cite{he2025cl}} & \multicolumn{1}{c}{87.47\tiny$\pm$0.60} & \multicolumn{1}{c}{91.37\tiny$\pm$1.30} &  \multicolumn{1}{c}{57.12\tiny$\pm$1.31} & \multicolumn{1}{c}{68.17\tiny$\pm$1.91} & \multicolumn{1}{c}{79.78\tiny$\pm$0.17} & \multicolumn{1}{c}{85.10\tiny$\pm$0.67} & \multicolumn{1}{c}{76.28\tiny$\pm$2.70} & \multicolumn{1}{c}{86.81\tiny$\pm$1.24} \\
    \multicolumn{1}{l}{LoRA-DRS~\cite{liu2025lora}} & \multicolumn{1}{c}{90.03\tiny$\pm$0.07} & \multicolumn{1}{c}{93.24\tiny$\pm$0.97} &  \multicolumn{1}{c}{57.58\tiny$\pm$0.79} & \multicolumn{1}{c}{66.72\tiny$\pm$0.79} & \multicolumn{1}{c}{75.96\tiny$\pm$0.36} & \multicolumn{1}{c}{81.82\tiny$\pm$0.85} & \multicolumn{1}{c}{87.58\tiny$\pm$0.28} & \multicolumn{1}{c}{92.30\tiny$\pm$0.61} \\
    \multicolumn{1}{l}{MOS~\cite{sun2025mos}} & \multicolumn{1}{c}{91.54\tiny$\pm$0.45} & \multicolumn{1}{c}{94.18\tiny$\pm$1.15} &  \multicolumn{1}{c}{57.54\tiny$\pm$0.37} & \multicolumn{1}{c}{64.50\tiny$\pm$1.44} & \multicolumn{1}{c}{77.65\tiny$\pm$0.50} & \multicolumn{1}{c}{81.94\tiny$\pm$0.66} & \multicolumn{1}{c}{89.88\tiny$\pm$0.29} & \multicolumn{1}{c}{93.52\tiny$\pm$0.61} \\
    \multicolumn{1}{l}{MACIL~\cite{wu2025navigating}} &
    \multicolumn{1}{c}{\underline{91.86\tiny$\pm$0.22}} &
    \multicolumn{1}{c}{\underline{94.44\tiny$\pm$0.96}} &
    \multicolumn{1}{c}{\underline{63.15\tiny$\pm$0.17}} &
    \multicolumn{1}{c}{\underline{70.54\tiny$\pm$1.79}} &
    \multicolumn{1}{c}{\textbf{81.82\tiny$\pm$0.22}} &
    \multicolumn{1}{c}{\underline{85.76\tiny$\pm$0.32}} &
    \multicolumn{1}{c}{\underline{90.23\tiny$\pm$0.13}} &
    \multicolumn{1}{c}{\underline{93.78\tiny$\pm$0.40}} \\
    \midrule
    \multicolumn{1}{l}{SR$^2$-LoRA} &
    \multicolumn{1}{c}{\textbf{91.98\tiny$\pm$0.10}} &
    \multicolumn{1}{c}{\textbf{94.60\tiny$\pm$0.08}} &
    \multicolumn{1}{c}{\textbf{63.27\tiny$\pm$0.78}} &
    \multicolumn{1}{c}{{70.03\tiny$\pm$1.02}} &
    \multicolumn{1}{c}{\underline{81.14\tiny$\pm$0.23}} &
    \multicolumn{1}{c}{\textbf{85.93\tiny$\pm$0.54}} &
    \multicolumn{1}{c}{\textbf{90.37\tiny$\pm$0.12}} &
    \multicolumn{1}{c}{\textbf{94.01\tiny$\pm$0.79}} \\
    \bottomrule
  \end{tabular}}
  \label{tab:four_benchmarks_results}
\end{table*}

\subsection{Ablation Study}

\textbf{Impact of Alignment Strategy.} Table~\ref{tab:alignment_20_50} reports an ablation study on different alignment strategies. We consider feature-level alignment on the final layer (Feature-L) and on all layers (Feature-A). For both variants, we also evaluate a setting without feature normalization to preserve the magnitude of layer-wise representations (w/o Norm). In addition, we include direct element-wise alignment of relation matrices (P2P), as well as two singular value variants, eigenvalue alignment on relation matrices averaged at the batch level (B-Eigen) and singular value alignment based on sample-level SVD (Eigen). The results show that preserving inter-layer relations consistently outperforms feature-level alignment, yielding better retention of previously learned knowledge. This observation is in line with Theorem~\ref{theorem_forgetting_relation}. Among relation-preserving strategies, Eigen achieves higher accuracy and lower variance than both P2P and B-Eigen. The comparable performance of P2P and B-Eigen suggests that averaging relation matrices across a batch weakens instance-specific structure. In contrast, sample-level singular value alignment retains such structure and exhibits improved stability, which is consistent with Proposition~\ref{prop:sv_stability}.

\begin{figure}[t]
\centering
\begin{minipage}{0.49\linewidth}
\centering
\includegraphics[width=0.98\linewidth,height=3.8cm]{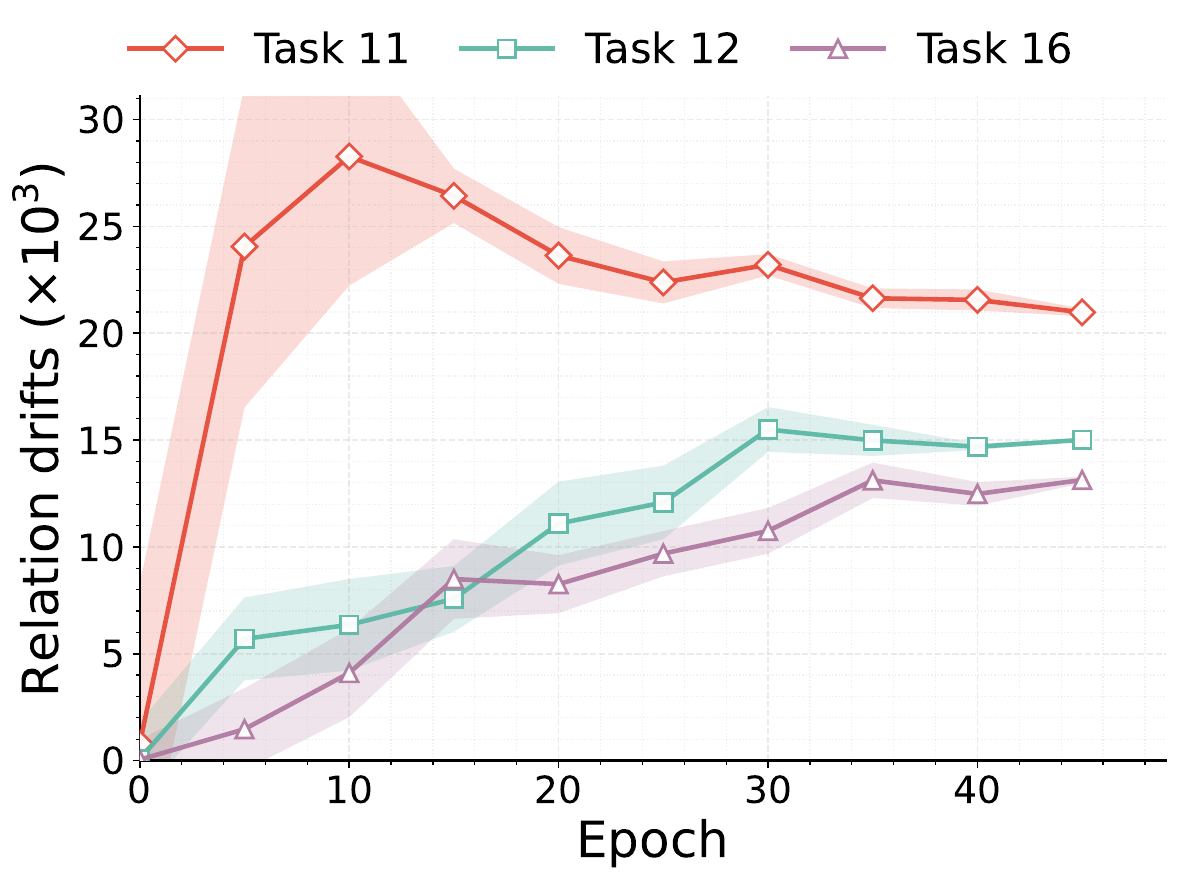}
\end{minipage}
\hfill
\begin{minipage}{0.49\linewidth}
\centering
\includegraphics[width=\linewidth,height=3.4cm]{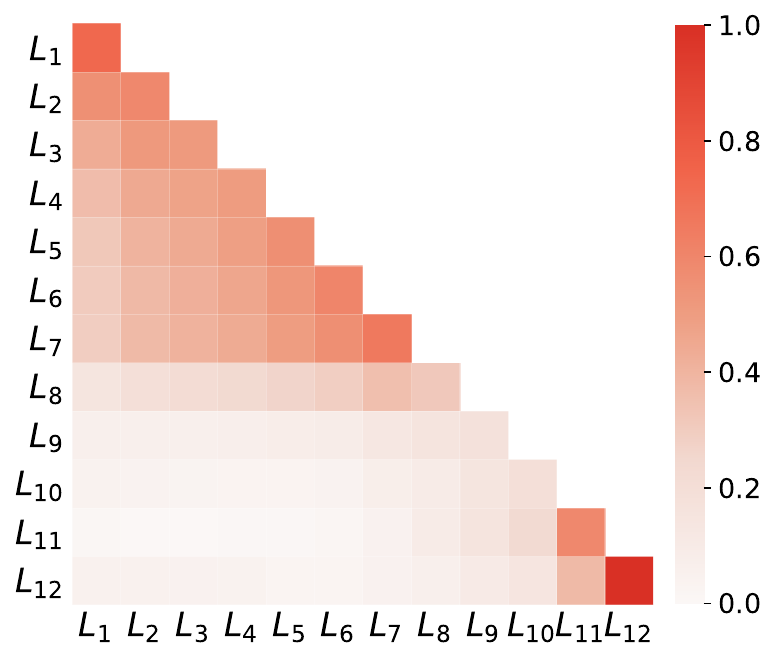}
\hfill
\end{minipage}
\caption{Visualization of inter-layer relation drift during training. (left) Evolution of inter-layer relation drift over training epochs for current-task samples. (right) Inter-layer relations after training, illustrating the layer-wise structural pattern.}
\label{fig:epoch_drift}
\end{figure}

\begin{table*}[t]
\caption{Ablation study of alignment strategies with 20 tasks. The mean and standard deviation over three runs are reported. All methods use the same ViT-B/16-IN21K backbone, random seeds, and class orders.}
\label{tab:alignment_20_50}
\centering
\resizebox{\textwidth}{!}{
\begin{tabular}{lcccccccc}
\toprule
\multirow{2}{*}{Alignment}&
\multicolumn{2}{c}{\textbf{CIFAR-100}} &
\multicolumn{2}{c}{\textbf{ImageNet-A}} &
\multicolumn{2}{c}{\textbf{ImageNet-R}} &
\multicolumn{2}{c}{\textbf{CUB-200}} \\
\cmidrule(lr){2-3} \cmidrule(lr){4-5} \cmidrule(lr){6-7} \cmidrule(l){8-9}
& $\mathcal{A}_{\mathrm{Last}}$ & $\mathcal{A}_{\mathrm{Avg}}$
& $\mathcal{A}_{\mathrm{Last}}$ & $\mathcal{A}_{\mathrm{Avg}}$ 
& $\mathcal{A}_{\mathrm{Last}}$ & $\mathcal{A}_{\mathrm{Avg}}$ 
& $\mathcal{A}_{\mathrm{Last}}$ & $\mathcal{A}_{\mathrm{Avg}}$ \\
\midrule
\multicolumn{1}{c}{Feature-L}
& 76.06{\tiny$\pm$1.95} & 86.22{\tiny$\pm$0.67}
& 35.64{\tiny$\pm$1.59} & 49.79{\tiny$\pm$0.30}
& 60.54{\tiny$\pm$0.75} & 71.51{\tiny$\pm$0.26}
& 57.56{\tiny$\pm$2.03} & 74.02{\tiny$\pm$2.94} \\
\multicolumn{1}{c}{w/o Norm}
& 89.68{\tiny$\pm$0.47} & 93.18{\tiny$\pm$0.34}
& 57.54{\tiny$\pm$1.44} & 66.17{\tiny$\pm$1.02}
& 78.25{\tiny$\pm$0.18} & 83.53{\tiny$\pm$0.34}
& 88.80{\tiny$\pm$0.15} & 93.39{\tiny$\pm$0.56} \\
\multicolumn{1}{c}{Feature-A}
& 75.57{\tiny$\pm$1.63} & 85.94{\tiny$\pm$0.69}
& 34.08{\tiny$\pm$1.94} & 49.47{\tiny$\pm$0.13}
& 60.55{\tiny$\pm$1.85} & 71.28{\tiny$\pm$0.38}
& 57.69{\tiny$\pm$2.29} & 73.96{\tiny$\pm$2.63} \\
\multicolumn{1}{c}{w/o Norm}
& 86.66{\tiny$\pm$0.20} & 92.84{\tiny$\pm$0.07}
& 53.80{\tiny$\pm$1.27} & 64.19{\tiny$\pm$0.56}
& 73.82{\tiny$\pm$0.78} & 81.42{\tiny$\pm$0.26}
& 81.62{\tiny$\pm$0.76} & 89.80{\tiny$\pm$1.03} \\
\hdashline
\multicolumn{1}{c}{P2P}
& 90.13{\tiny$\pm$0.37} & 93.94{\tiny$\pm$0.33}
& 59.15{\tiny$\pm$0.91} & 68.47{\tiny$\pm$0.92}
& 79.01{\tiny$\pm$0.27} & 84.19{\tiny$\pm$0.56}
& 89.75{\tiny$\pm$0.32} & 93.69{\tiny$\pm$0.89} \\
\multicolumn{1}{c}{B-Eigen}
& 90.18{\tiny$\pm$0.40} & 93.97{\tiny$\pm$0.36}
& 59.57{\tiny$\pm$1.78} & 68.53{\tiny$\pm$1.02}
& 79.26{\tiny$\pm$0.41} & 84.13{\tiny$\pm$0.66}
& 89.80{\tiny$\pm$0.27} & 93.70{\tiny$\pm$0.91} \\
\multicolumn{1}{c}{Eigen}
& 90.45{\tiny$\pm$0.14} & 94.08{\tiny$\pm$0.12}
& 61.01{\tiny$\pm$0.79} & 69.41{\tiny$\pm$0.12}
& 79.54{\tiny$\pm$0.45} & 84.32{\tiny$\pm$0.53}
& 89.71{\tiny$\pm$0.21} & 94.32{\tiny$\pm$0.11} \\
\bottomrule
\end{tabular}
}
\end{table*}

\begin{figure}[t]
\centering
\begin{minipage}{0.49\linewidth}
\centering
\includegraphics[width=\linewidth]{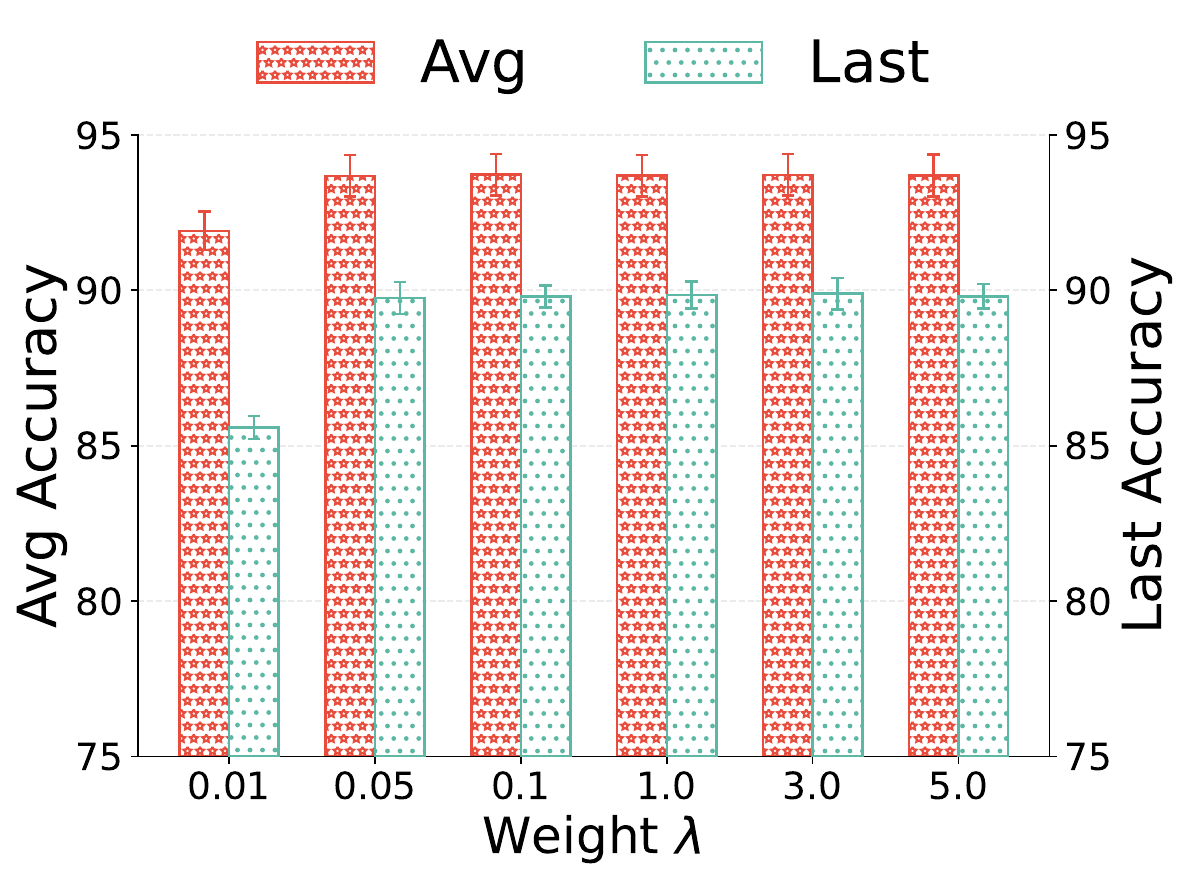}
\\(a) Constant $\lambda$
\end{minipage}
\hfill
\begin{minipage}{0.49\linewidth}
\centering
\includegraphics[width=\linewidth]{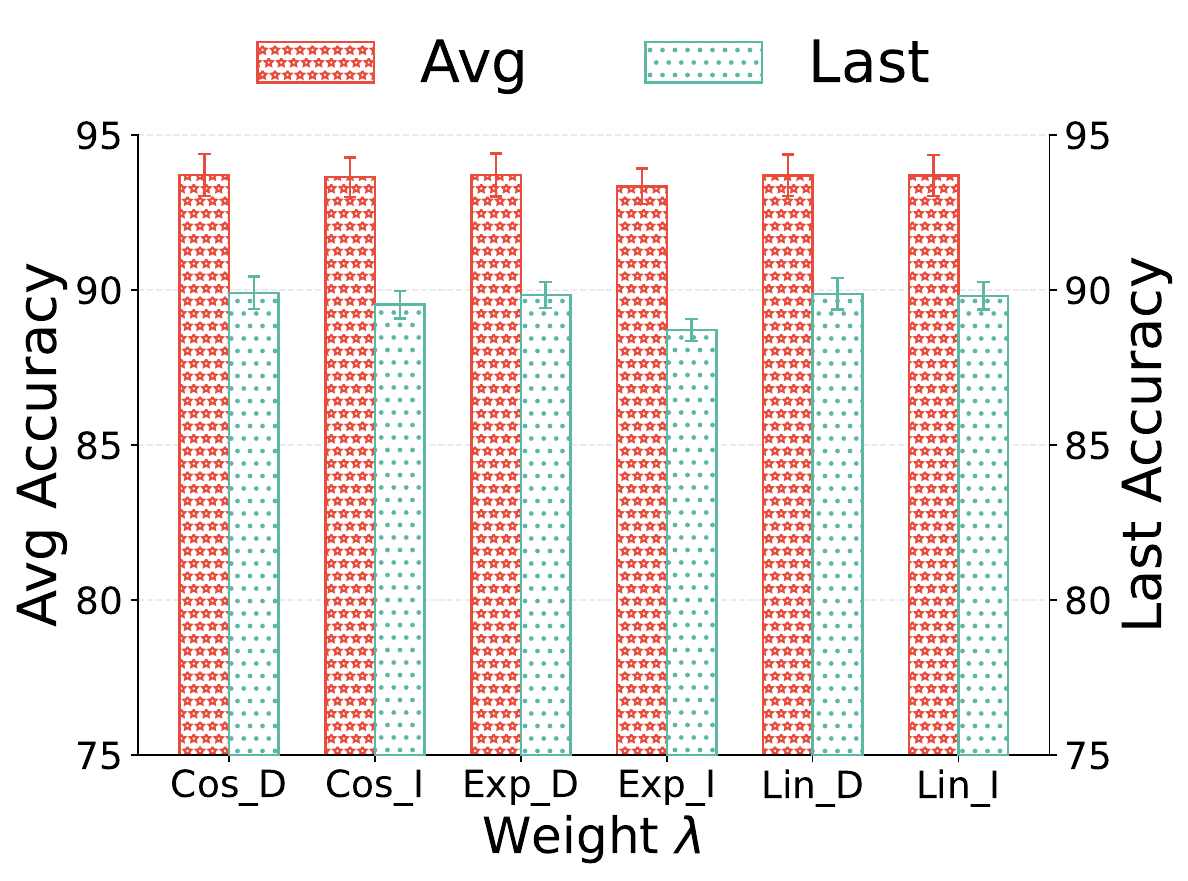}
\\(b) Dynamic $\lambda$
\end{minipage}
\caption{Comparison of constant and dynamically scheduled trade-off coefficients \(\lambda\) under the 20-task setting on CUB200.}
\label{fig:lambd}
\end{figure}

\textbf{Effect of Adjustment Strategies for $\lambda$.} Fig.~\ref{fig:lambd} analyzes the effect of different scheduling strategies for the trade-off coefficient \(\lambda\) under the 20-task setting, which is representative of long task sequences where cumulative drift becomes more pronounced. Considering that inter-layer relation drift evolves throughout training, as illustrated in Fig.~\ref{fig:epoch_drift}~(a), we evaluate a constant coefficient alongside several dynamic schedules, including cosine, exponential, and linear variants. Under a constant coefficient, performance remains stable across a broad range of \(\lambda\) values and deteriorates only when the alignment constraint becomes negligible. This trend indicates that insufficient regularization fails to effectively constrain inter-layer relation drift, thereby increasing the disruption of previously established relations and leading to more severe forgetting. In contrast, dynamic scheduling strategies, although designed to adapt to the temporal evolution of relation drift, do not provide noticeable improvements over a fixed coefficient. This observation suggests that the proposed singular value alignment objective is well-conditioned and inherently robust to the choice of \(\lambda\), enabling stable control of inter-layer relation drift without requiring carefully tuned schemes.

\begin{figure}[t]
\centering
\begin{minipage}{0.49\linewidth}
\centering
\includegraphics[width=\linewidth]{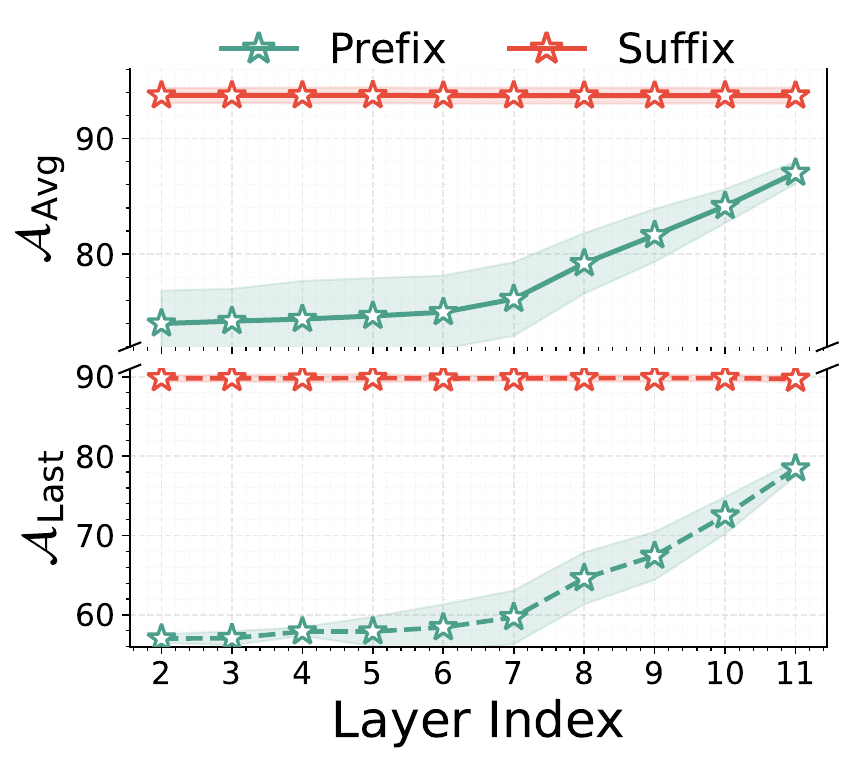}
\\(a) CUB-200
\end{minipage}
\hfill
\begin{minipage}{0.49\linewidth}
\centering
\includegraphics[width=\linewidth]{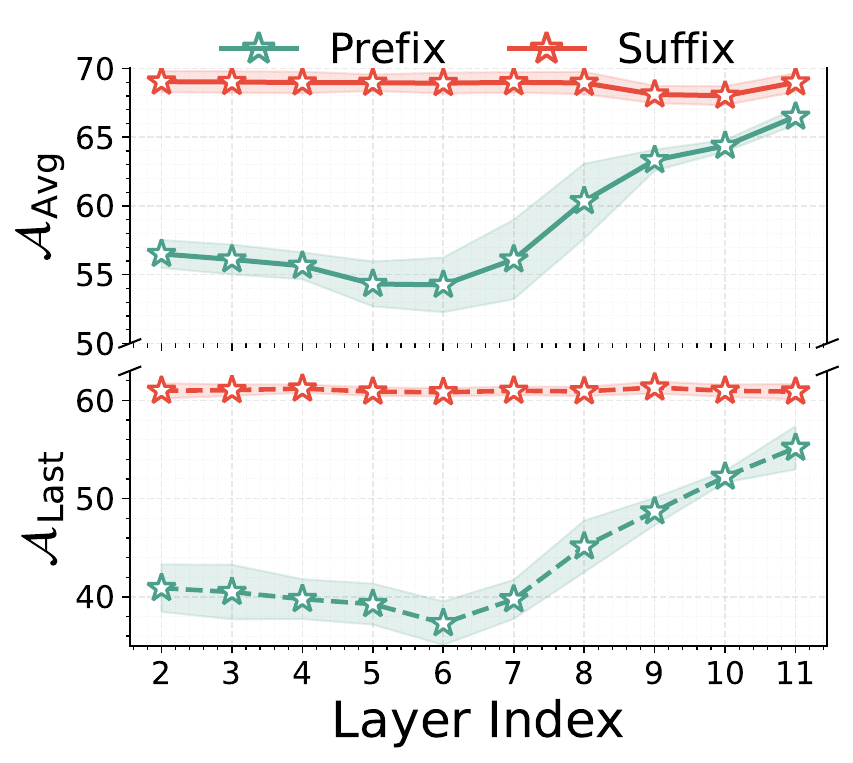}
\\(b) ImageNet-A
\end{minipage}
\caption{Effect of alignment depth on performance across datasets under the 20-task setting. Prefix and Suffix denote shallow-to-deep and deep-to-shallow alignment, respectively.}
\label{fig:layer}
\end{figure}

\subsection{Further Analysis}

\begin{table*}[t]
\caption{Performance under different LoRA configurations $r$ under the 20-task setting on CUB200. The mean and standard deviation over three runs are reported. All methods use the same ViT-B/16-IN21K backbone, random seeds, and class orders.}
\centering
\resizebox{\textwidth}{!}{%
\begin{tabular}{lcccccccc}
\toprule
\multirow{2}{*}{\textbf{LoRA Config}}
& \multicolumn{2}{c}{$r=1$}
& \multicolumn{2}{c}{$r=10$}
& \multicolumn{2}{c}{$r=20$}
& \multicolumn{2}{c}{$r=64$} \\
\cmidrule(lr){2-3}\cmidrule(lr){4-5}\cmidrule(lr){6-7}\cmidrule(lr){8-9}
& $\mathcal{A}_{\mathrm{Last}}$ & $\mathcal{A}_{\mathrm{Avg}}$
& $\mathcal{A}_{\mathrm{Last}}$ & $\mathcal{A}_{\mathrm{Avg}}$
& $\mathcal{A}_{\mathrm{Last}}$ & $\mathcal{A}_{\mathrm{Avg}}$
& $\mathcal{A}_{\mathrm{Last}}$ & $\mathcal{A}_{\mathrm{Avg}}$ \\
\midrule
$\{W_v\}$
& 93.91{\tiny$\pm$0.72} & 89.16{\tiny$\pm$0.28}
& 93.93{\tiny$\pm$0.94} & 89.21{\tiny$\pm$0.67}
& 93.92{\tiny$\pm$0.62} & 89.58{\tiny$\pm$0.31}
& 93.99{\tiny$\pm$0.69} & 89.03{\tiny$\pm$0.30} \\
$\{W_k, W_v\}$
& 93.96{\tiny$\pm$0.64} & 89.41{\tiny$\pm$0.35}
& 94.32{\tiny$\pm$0.69} & 89.71{\tiny$\pm$0.21}
& 94.05{\tiny$\pm$0.84} & 89.13{\tiny$\pm$0.11}
& 94.14{\tiny$\pm$0.75} & 89.46{\tiny$\pm$0.43} \\
$\{W_q, W_v\}$
& 93.92{\tiny$\pm$0.76} & 89.45{\tiny$\pm$0.40}
& 94.08{\tiny$\pm$0.63} & 89.63{\tiny$\pm$0.39}
& 93.95{\tiny$\pm$0.59} & 89.06{\tiny$\pm$0.35}
& 94.17{\tiny$\pm$0.27} & 89.22{\tiny$\pm$0.48} \\
$\{W_q, W_k, W_v\}$
& 93.98{\tiny$\pm$0.69} & 89.64{\tiny$\pm$0.54}
& 93.97{\tiny$\pm$0.74} & 89.67{\tiny$\pm$0.58}
& 93.93{\tiny$\pm$0.69} & 89.34{\tiny$\pm$0.71}
& 94.06{\tiny$\pm$0.97} & 89.65{\tiny$\pm$0.50} \\

\bottomrule
\end{tabular}%
}
\label{tab:rank_r_results}
\end{table*}

\textbf{Impact of LoRA Configurations.} We examine different LoRA configurations by varying the adaptation rank \(r\) and the choice of projection matrices \((W_q, W_k, W_v)\), following prior work~\cite{he2025cl,liang2024inflora}. Results on CUB-200 with 20 tasks are reported in Table~\ref{tab:rank_r_results}. The configuration \(\{W_k, W_v\}\) with \(r=10\) achieves the best overall performance. This suggests that jointly adapting the key and value projections provides a more suitable parameterization for controlling inter-layer relation drift, as it better preserves the structural consistency of layer-wise representations during continual adaptation. The effect of the adaptation rank is not monotonic. Moderate ranks generally yield stronger performance, while increasing the rank does not bring further improvement. This observation suggests that a moderate adaptation capacity is sufficient to capture the changes required by new tasks while maintaining stable inter-layer relations.

\begin{table}[t]
\centering
\caption{Computational overhead comparison under the 20-task setting. LP (M), TT (s), and IT (s) denote learnable parameters, training time, and inference time, respectively. Inference time and training time are measured at the final task.}
\label{tab:efficiency}
\resizebox{\linewidth}{!}{
\begin{tabular}{lccccc}
\toprule
Method & LP & TT & IT & $\mathcal{A}_{\mathrm{Last}}$  \\
\midrule
Vanilla  & 0.38 & 131.10 & 6.20 & 47.06{\tiny$\pm$4.90} \\
CODA-Prompt  & 3.99 & 90.49 & 7.48 & 66.41{\tiny$\pm$0.81} \\
SLCA & 85.81 & 139.20 & 4.42 & 82.48{\tiny$\pm$0.53} \\
EASE & 0.76 & 78.88 & 62.46 & 84.51{\tiny$\pm$1.67} \\
MOS  & 0.76 & 60.74 & 66.95 & 89.42{\tiny$\pm$0.22}\\
LoRA-DRS  & 0.38 & 178.74 & 6.20 & 87.50{\tiny$\pm$0.26} \\
MACIL  & 1.19 & 1351.25 & 6.20 & 88.63{\tiny$\pm$0.61} \\
\midrule
SR$^2$-LoRA & 0.38 & 137.01 & 6.20 &89.71{\tiny$\pm$0.21} \\
\bottomrule
\end{tabular}
}
\end{table}

\textbf{Impact of Alignment Depth.} Fig.~\ref{fig:layer} compares Prefix and Suffix alignment strategies, corresponding to shallow-to-deep and deep-to-shallow alignment, respectively. Performance increases as more layers are aligned, with the best results achieved when all layers are included. Under partial alignment, Suffix consistently outperforms Prefix and attains competitive performance with only a small subset of deepest layers, whereas Prefix requires substantially more layers to reach a similar level. These observations indicate that inter-layer relation drift in deeper layers exerts a more direct influence on forgetting. Deeper layers are more closely tied to task-specific representations and decision boundaries, so perturbations in their inter-layer relations are more likely to affect classification margins. In contrast, shallow layers capture more general representations and exhibit more redundant relation structures. This interpretation is supported by Fig.~\ref{fig:epoch_drift}~(right), where shallow layers show denser relation patterns while deeper layers exhibit sparser and more selective dependencies. 

\begin{figure}[t]
\centering
\begin{minipage}{\linewidth}
\centering
\includegraphics[width=0.49\linewidth]{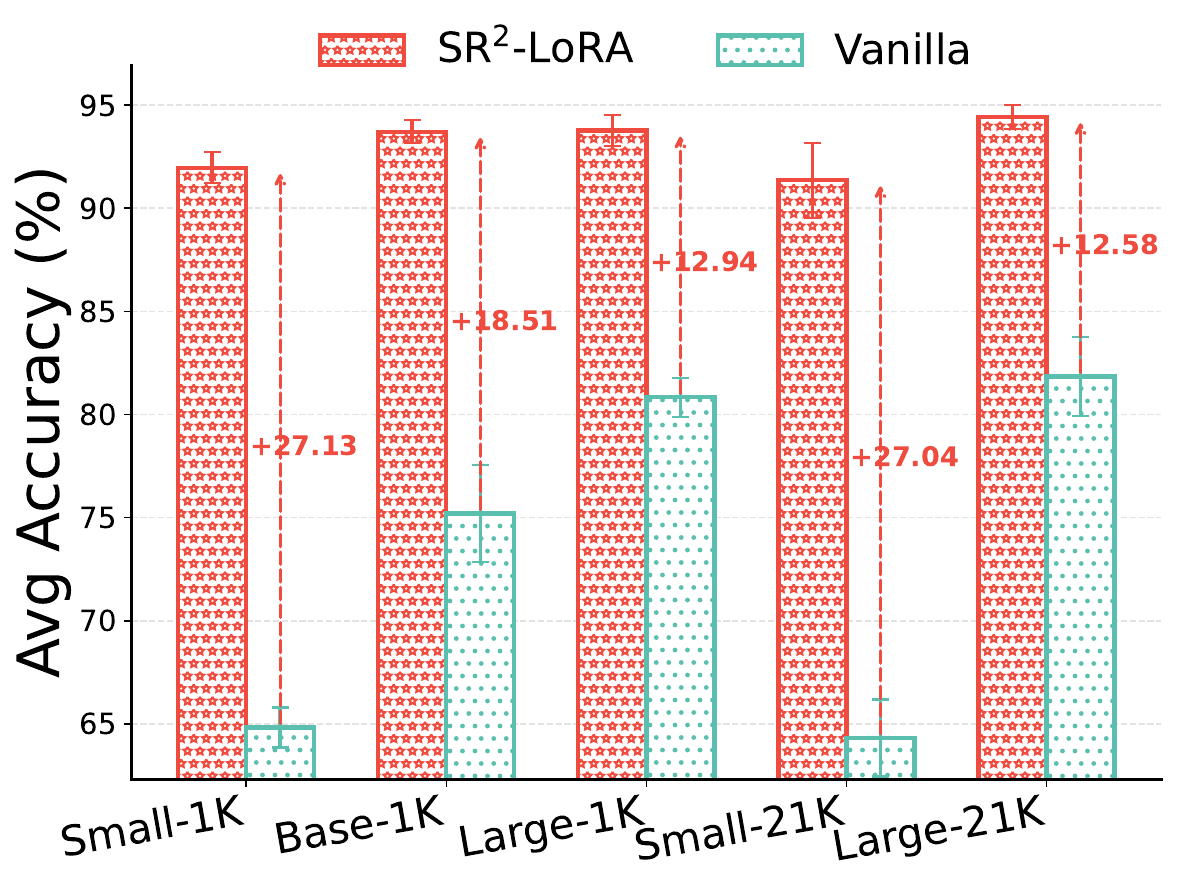}
\includegraphics[width=0.49\linewidth]{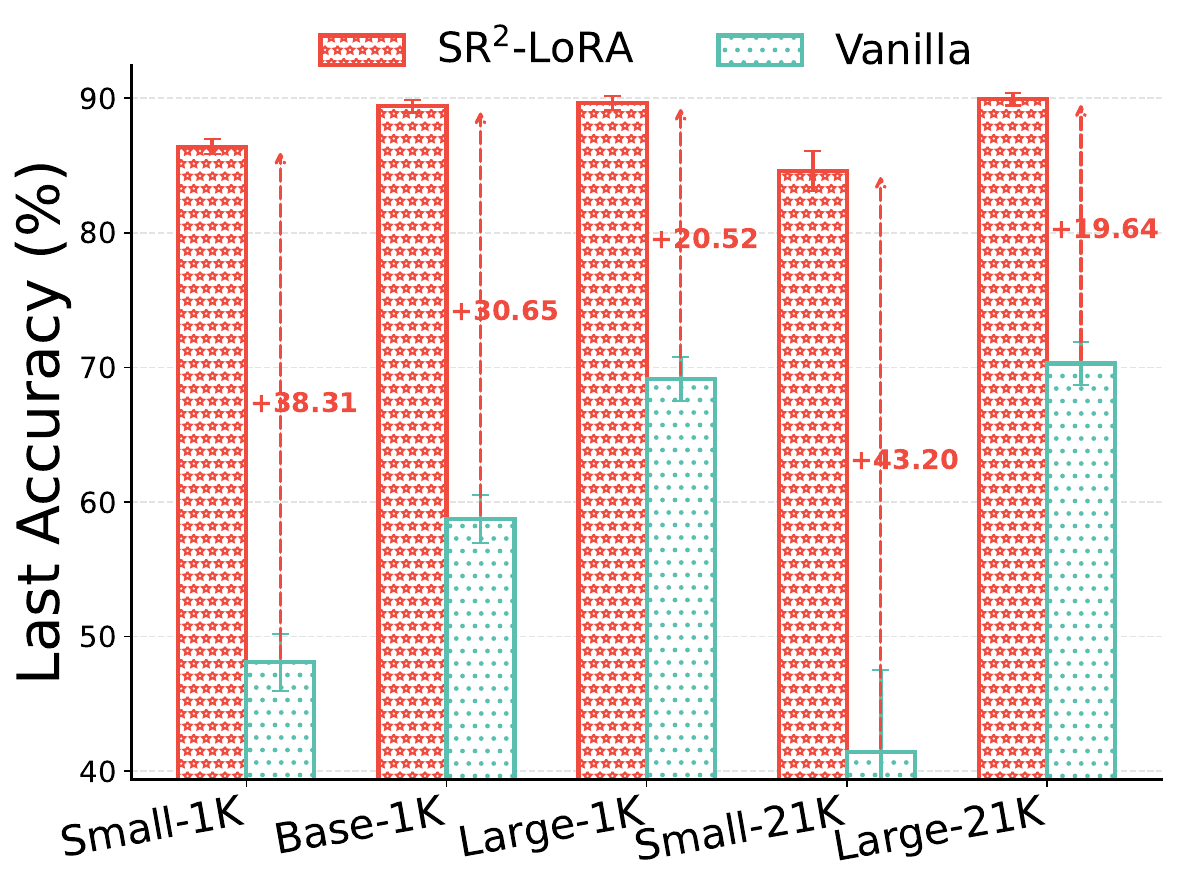}
\\(a) CUB-200
\end{minipage}
\hfill
\begin{minipage}{\linewidth}
\centering
\includegraphics[width=0.49\linewidth]{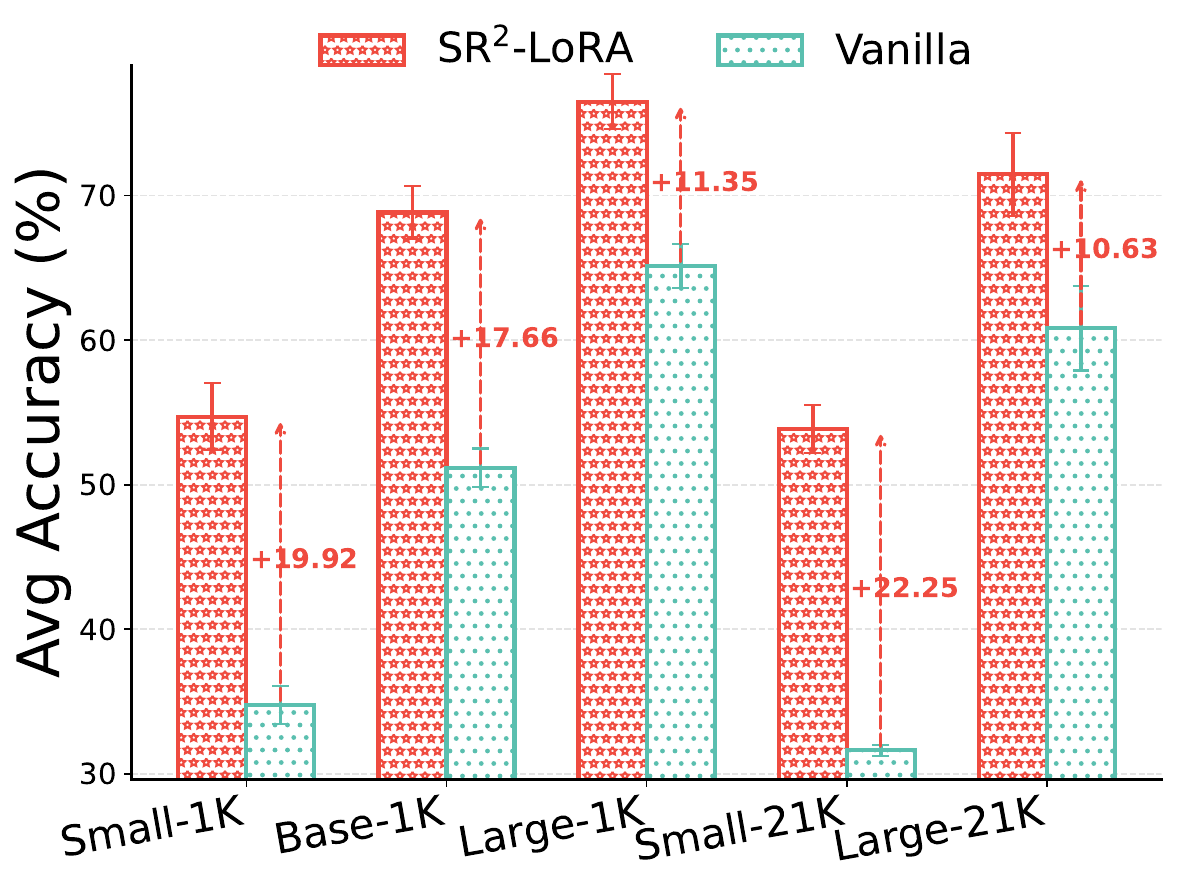}
\includegraphics[width=0.49\linewidth]{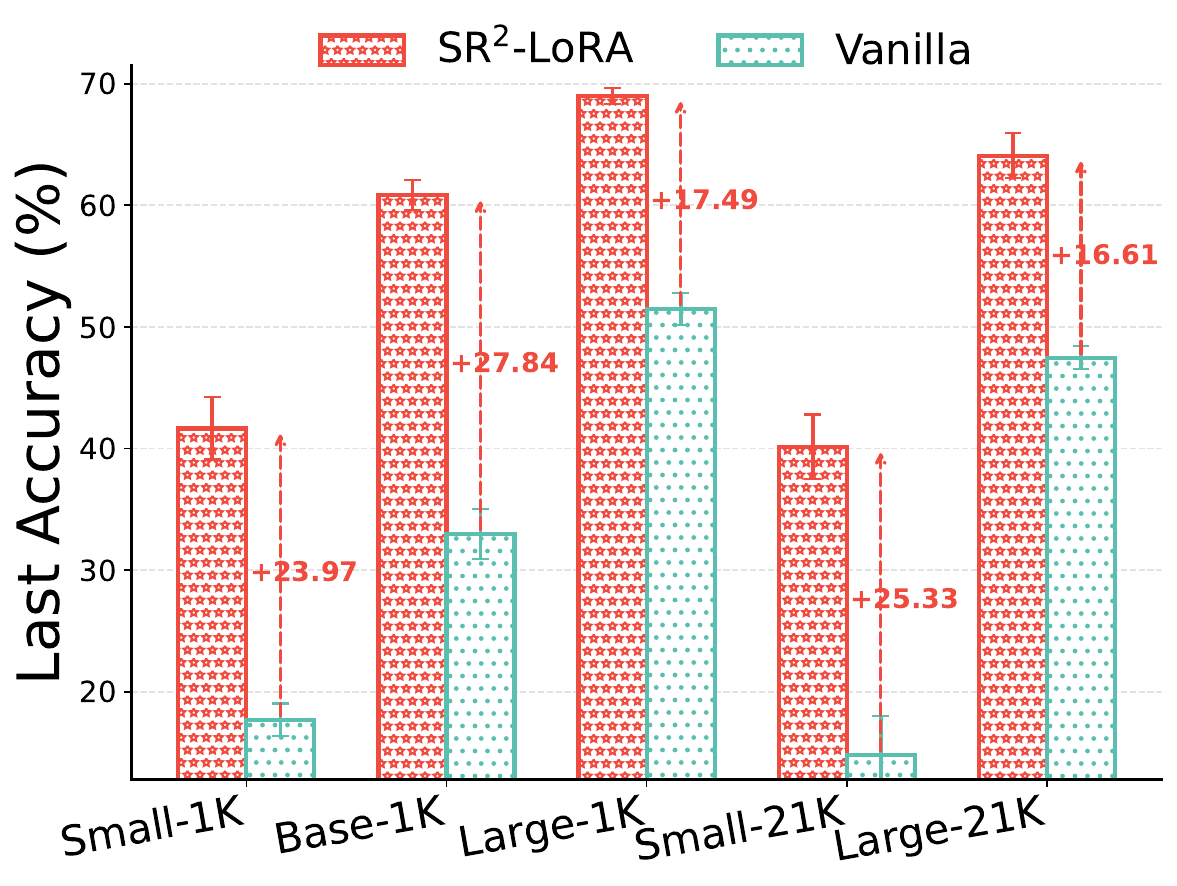}
\\(b) ImageNet-A
\end{minipage}
\caption{Effect of different PTMs on performance under the 20-task setting. Results on (a) CUB-200 and (b) ImageNet-A compare Vanilla and SR$^2$-LoRA across ViT with different depths and pre-training datasets.}
\label{fig:backbone}
\end{figure}

\textbf{Computational Overhead.} As shown in Table~\ref{tab:efficiency}, SR$^2$-LoRA introduces negligible computational overhead compared to vanilla training, with both the learnable parameters and the runtime, including training and inference time, remaining comparable. From a complexity perspective, the additional cost arises from the construction of inter-layer relation matrices and the SVD in the alignment objective, which incur $\mathcal{O}(L^2 d)$ and $\mathcal{O}(L^3)$ complexity, respectively, where $L$ denotes the number of layers and $d$ the feature dimension. Since the number of layers is small, with $L=12$ in ViT-B/16, these costs are negligible compared to the backbone forward computation and are incurred only during training. This efficiency makes SR$^2$-LoRA well-suited for long task sequences.

\textbf{Impact of Different PTMs.} To evaluate the generality of the proposed method, we examine SR$^2$-LoRA across Vision Transformers with different depths and pre-training schemes, as shown in Fig.~\ref{fig:backbone}. Experiments are conducted on CUB200 and ImageNet-A under the 20-task setting, including ViT-S and ViT-B with 12 layers and ViT-L with 24 layers, as well as models pre-trained on ImageNet-1K and ImageNet-21K. SR$^2$-LoRA consistently improves performance across all configurations. The gains are more pronounced for deeper models, where larger capacity leads to stronger adaptation to new tasks and thus increased susceptibility to forgetting. By constraining inter-layer relations, the proposed method effectively stabilizes the representation structure, mitigating this effect. In addition, comparable improvements are observed under different pre-training datasets, indicating that the effectiveness of relation alignment is not dependent on the pre-training scale. These results demonstrate that the proposed method is robust to variations in backbone architecture and pre-training strategy.

\subsection{Visualization}

\begin{figure}[t]
\centering
\setlength{\tabcolsep}{1pt}
\begin{tabular}{c c c c}
 & \multicolumn{1}{c}{Task1} & \multicolumn{1}{c}{Task10} & \multicolumn{1}{c}{Task19} \\
\rotatebox{90}{\textbf{Vanilla}} &
\includegraphics[width=0.28\linewidth]{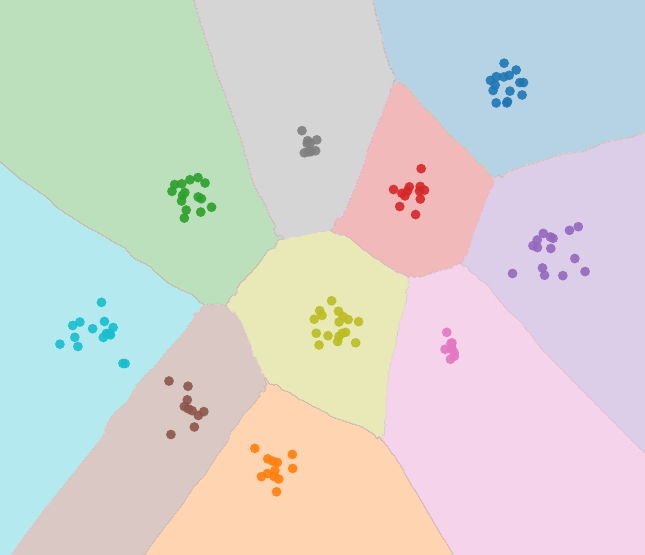} &
\includegraphics[width=0.28\linewidth]{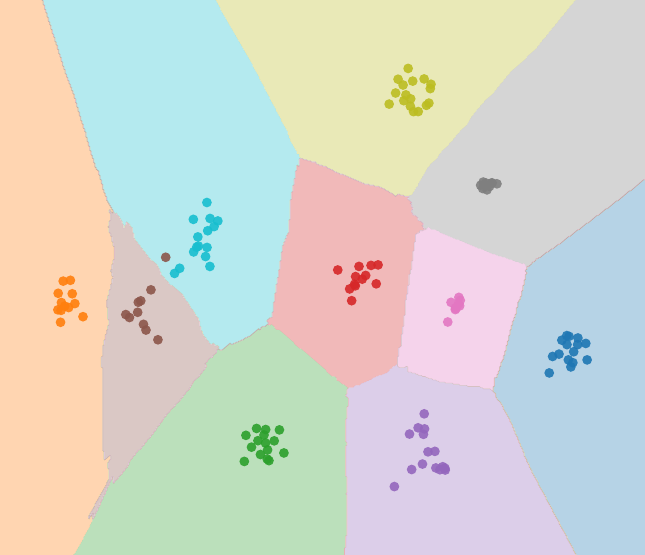} &
\includegraphics[width=0.28\linewidth]{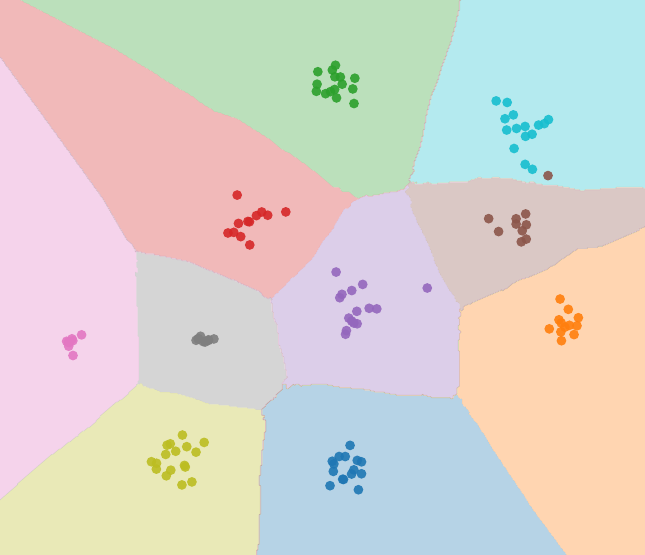} \\
\rotatebox{90}{\textbf{SR$^2$-LoRA}} &
\includegraphics[width=0.28\linewidth]{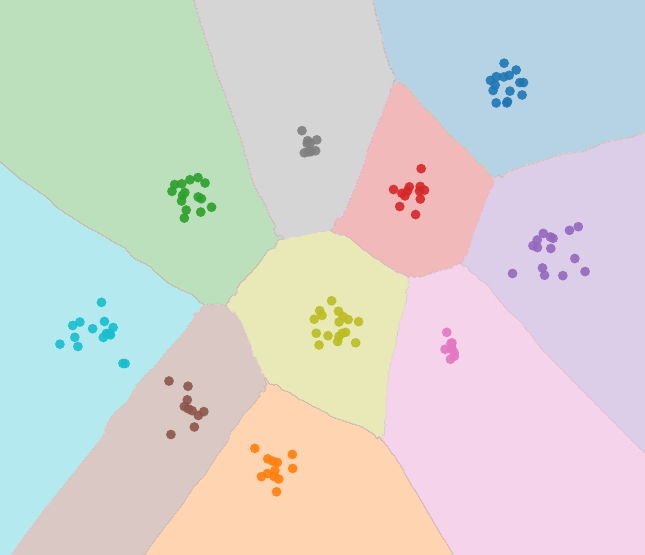} &
\includegraphics[width=0.28\linewidth]{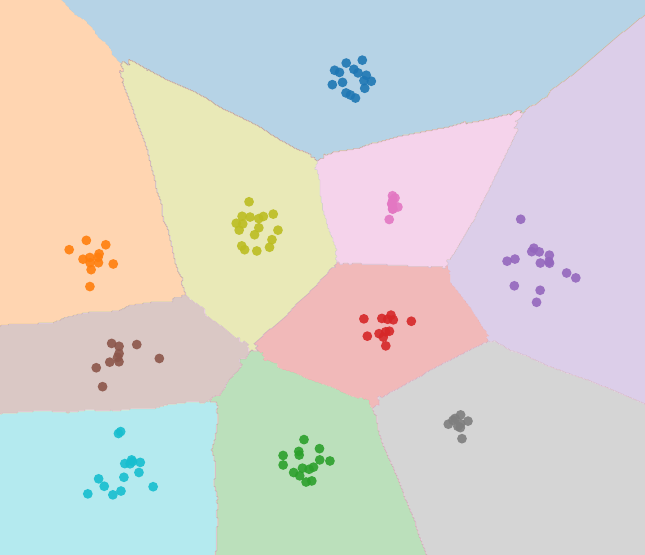} &
\includegraphics[width=0.28\linewidth]{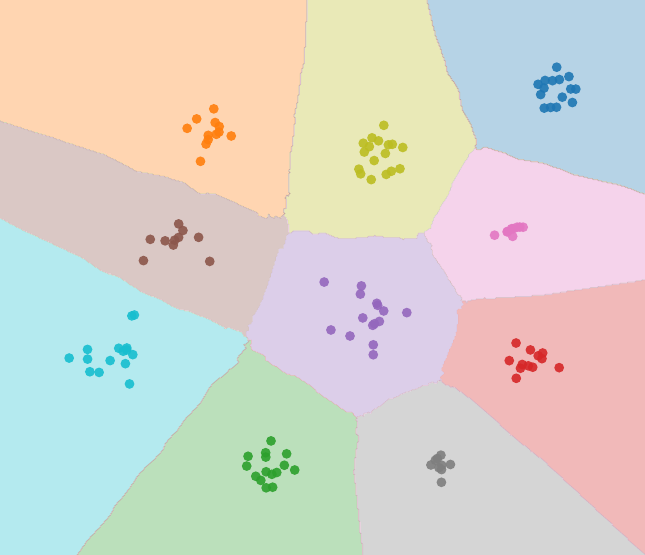} \\
\end{tabular}
\caption{Visualization of decision regions for Task 1 on CUB200 under the 20-task setting after learning Task 10 and Task 19. Decision regions are illustrated by the shaded background.}
\label{fig:tsne_comparison}
\end{figure}

\textbf{Decision Regions.} Fig.~\ref{fig:tsne_comparison} visualizes the decision regions for Task~1 to examine the impact of inter-layer relation drift on classification margins~\cite{2008Visualizing}. Under vanilla training, unconstrained relation drift progressively degrades classification margins, leading to substantial changes in the spatial organization of decision regions. As the model adapts to new tasks, the regions associated with Task~1 become increasingly less separable and exhibit clear structural disruption by Task~19, reflecting the effect of catastrophic forgetting. In contrast, SR$^2$-LoRA maintains remarkable stability. By constraining inter-layer relation drift, the proposed method preserves the relational structure established for previous tasks. As a result, the decision regions for Task~1 remain well-formed even after the full 20-task sequence. These observations provide visual evidence that controlling inter-layer relation drift is an effective mechanism for mitigating catastrophic forgetting.

\begin{figure}[t]
\centering
\begin{minipage}{0.49\linewidth}
\centering
\includegraphics[width=\linewidth]{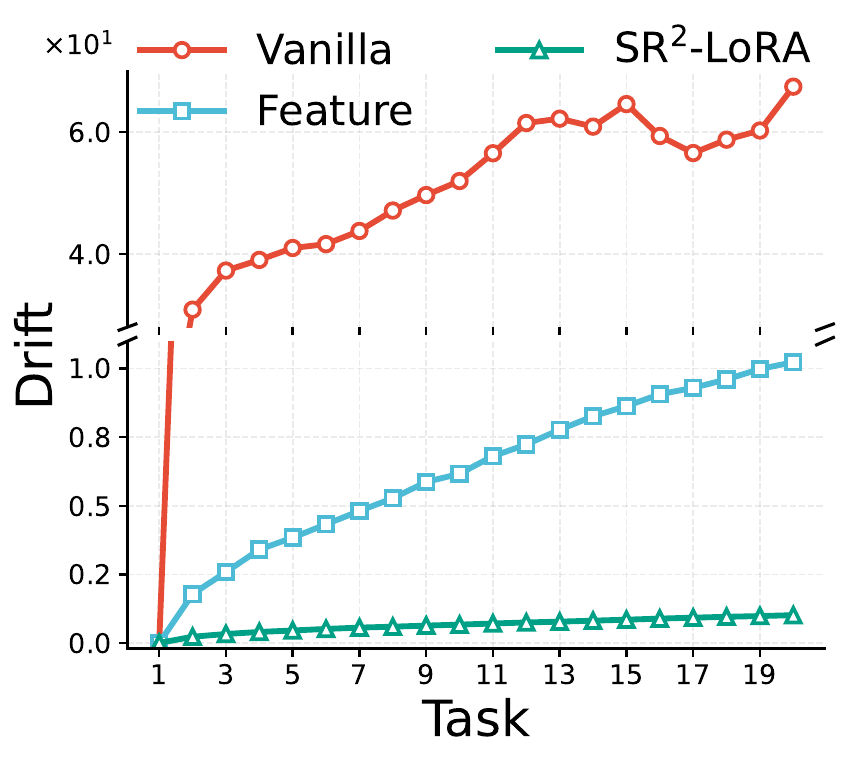}
\\(a) Feature Drift
\end{minipage}
\hfill
\begin{minipage}{0.49\linewidth}
\centering
\includegraphics[width=\linewidth]{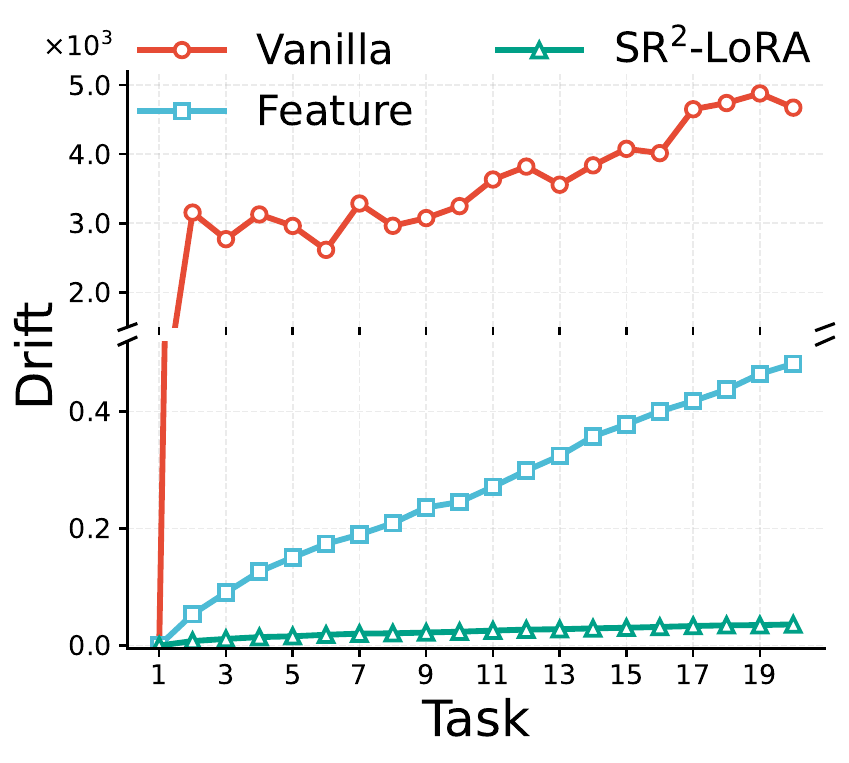}
\\(b) Relation Drift
\end{minipage}
\caption{Visualization of feature and relation drift on Task~1 on CUB200 under the 20-task setting for vanilla training, last-layer feature distillation, and SR$^2$-LoRA.}
\label{fig:drift}
\end{figure}

\textbf{Drift.} We further visualize feature drift and relation drift on Task~1 across the 20-task sequence, as shown in Fig.~\ref{fig:drift}. Consistent with Table~\ref{tab:alignment_20_50}, reducing representation drift alleviates forgetting compared with vanilla training. However, last-layer feature distillation and relation-level alignment show different behaviors. Last-layer feature distillation reduces feature drift, but relation drift still accumulates along the task sequence, indicating that constraining only the final representation is insufficient to preserve inter-layer relations. In contrast, SR$^2$-LoRA maintains substantially lower relation drift by directly constraining inter-layer relations. Meanwhile, the feature drift under SR$^2$-LoRA also remains limited, showing that relation-level alignment provides a stronger constraint on representation changes. These results further support the motivation of SR$^2$-LoRA. Directly constraining inter-layer relation drift better preserves the representation structure of previous-task samples than relying solely on last-layer feature distillation.

\section{Conclusion}

In this paper, we study catastrophic forgetting from a new perspective, namely inter-layer relation drift, and provide theoretical analysis showing that such drift plays a fundamental role in forgetting. From this perspective, we characterize forgetting as the accumulation of inter-layer relation drift, which motivates directly constraining inter-layer relations rather than focusing on individual layers. Based on this insight, we propose SR$^2$-LoRA, which constrains relation drift via singular value alignment, yielding a stable and well-conditioned objective for preserving inter-layer relations. Extensive experiments demonstrate that the proposed method effectively mitigates forgetting and consistently improves performance across datasets, model scales, and pre-training schemes. Moreover, SR$^2$-LoRA introduces negligible computational overhead.

\section*{Acknowledgment}

This work was supported in part by the NSFC~(62276131), in part by the Natural Science Foundation of Jiangsu Province of China under Grant (BK20240081).

\bibliographystyle{IEEEtran}
\bibliography{IEEEabrv,main}

\begin{IEEEbiography}
[{\includegraphics[width=1in,height=1.25in,clip,keepaspectratio]{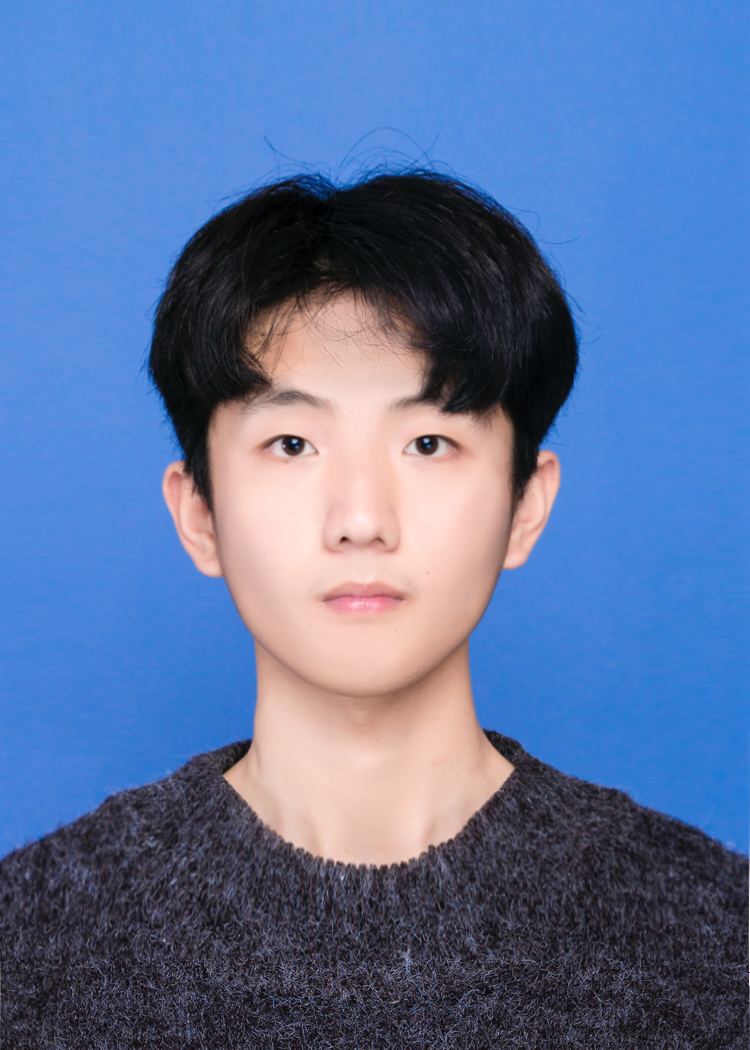}}]{Fengqiang Wan}
is currently working towards the Ph.D. degree at the School of Computer Science and Engineering, Nanjing University of Science and Technology. His research interests mainly lie in deep learning and data mining. He is currently focusing on incremental learning.
\end{IEEEbiography}

\begin{IEEEbiography}
[{\includegraphics[width=1in,height=1.25in,clip,keepaspectratio]{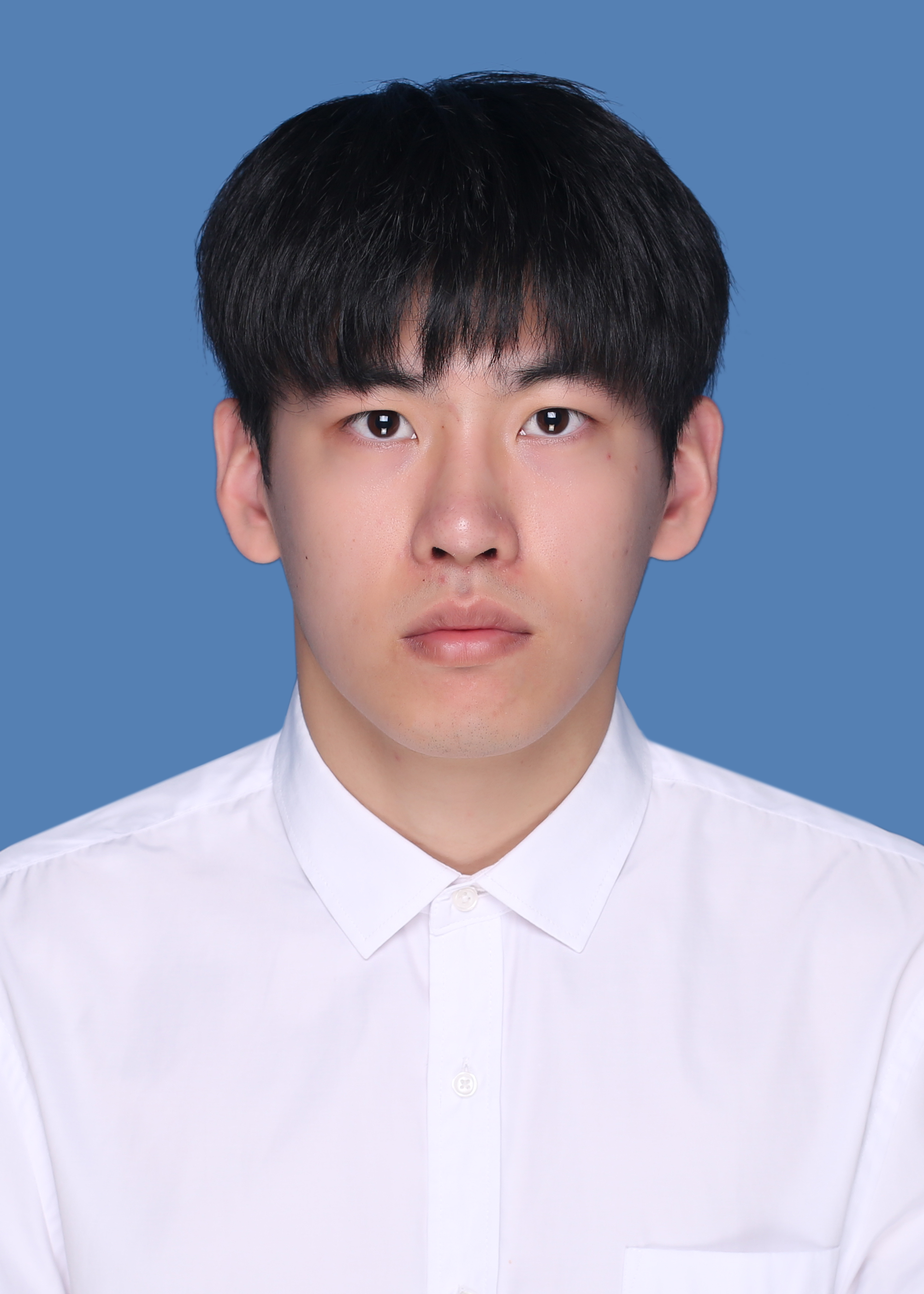}}]{Yipeng Lin}
is currently working towards the M.S. degree at the School of Computer Science and Engineering, Nanjing University of Science and Technology. His research interests mainly lie in deep learning and data mining. He is currently focusing on incremental learning.
\end{IEEEbiography}

\begin{IEEEbiography}
[{\includegraphics[width=1in,height=1.25in,clip,keepaspectratio]{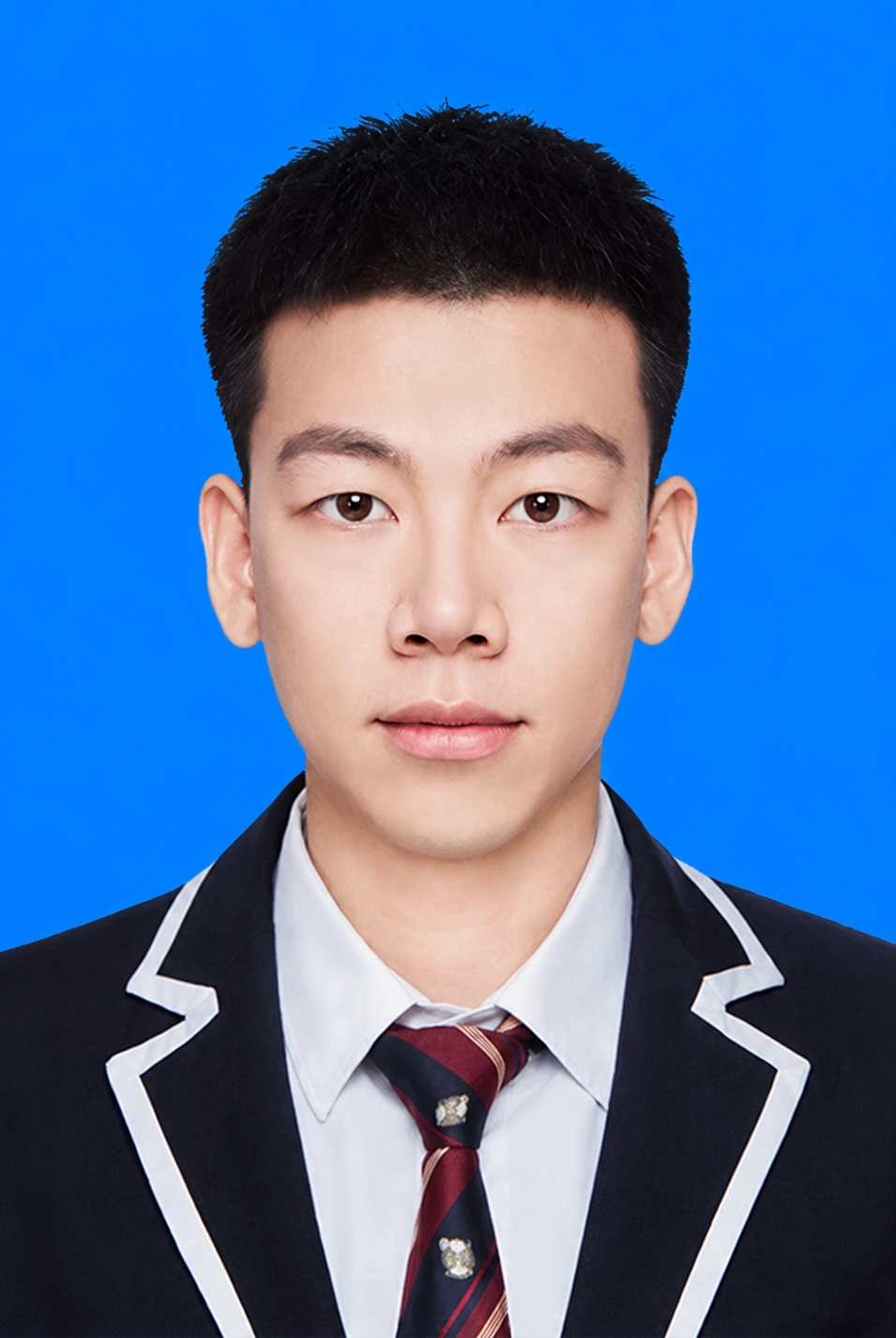}}]{Kan Lv}
is currently working towards the M.S. degree at the School of Computer Science and Engineering, Nanjing University of Science and Technology. His research interests mainly lie in deep learning and data mining. He is currently focusing on incremental learning.
\end{IEEEbiography}

\begin{IEEEbiography}
[{\includegraphics[width=1in,height=1.25in,clip,keepaspectratio]{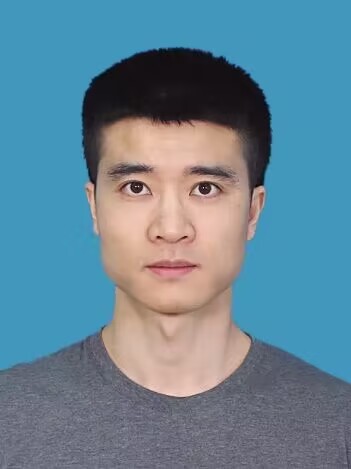}}]{Yang Yang}
received the Ph.D. degree in computer science, Nanjing University, China in 2019. At the same year, he became a faculty member at Nanjing University of Science and Technology, China. He is currently a Professor with the school of Computer Science and Engineering. His research interests lie primarily in machine learning and data mining, including heterogeneous learning, model reuse, and incremental mining.  He has published prolifically in refereed journals and conference proceedings, including IEEE Transactions on Knowledge and Data Engineering (TKDE), ACM Transactions on Information Systems (ACM TOIS), ACM Transactions on Knowledge Discovery from Data (TKDD), ACM SIGKDD, ACM SIGIR, WWW, IJCAI, and AAAI. He was the recipient of the the Best Paper Award of ACML-2017. He serves as PC in leading conferences such as IJCAI, AAAI, ICML, NeurIPS, etc.
\end{IEEEbiography}

\end{document}